\documentclass[conference]{IEEEtran}   	
\usepackage{geometry}                		
\usepackage[parfill]{parskip}    		        
\usepackage{graphicx}				
\usepackage{amssymb}				
\usepackage{amsmath}
\usepackage{textcomp}
\usepackage{breqn}
\usepackage{bbm}
\usepackage{tabularx}
\usepackage{latexsym}
\usepackage{subcaption}
\usepackage{gensymb}
\usepackage{amsmath}
\usepackage{fancyhdr}
\usepackage{eucal}
\usepackage{float}
\usepackage{parskip}
\usepackage{textcomp}

\usepackage{geometry}                		
\usepackage{tabularx}
\usepackage{stmaryrd}
\usepackage{wasysym} 
\usepackage{xurl}
\usepackage{multirow}
\usepackage[table]{xcolor} 
\newcommand{\code}[1]{\texttt{#1}}

\usepackage{microtype}
\usepackage{yfonts}
\usepackage{listings} 
\usepackage{multirow}
\usepackage{xcolor}
\usepackage{pdfpages}
\usepackage{graphicx} 
\usepackage[utf8]{inputenc}
\usepackage [english]{babel}
\usepackage [autostyle, english = american]{csquotes}
\usepackage{biblatex}
\graphicspath{{./Images/}} 
\usepackage{mathtools}
\usepackage{amsmath}
\usepackage{fancyhdr} 
\usepackage{titlesec}
\usepackage{upgreek} 
\usepackage{amsmath} 
\usepackage{amssymb}

\PassOptionsToPackage{hyphens, spaces, obeyspaces}{url} 
\usepackage[final=true,backref=true,colorlinks=true,linkcolor=blue, urlcolor=blue, citecolor=blue, breaklinks=true]{hyperref}
\hypersetup{
    pagebackref=true,
    colorlinks=true,
    citecolor=blue,
    linkcolor=blue,
    filecolor=magenta,      
    urlcolor=blue,
}
\hypersetup{breaklinks=true}
\usepackage{breakurl}

\def\xHyphenate#1#2\wholeString {\if#1$%
    \else\transform{#1}%
    \takeTheRest#2\ofTheString\fi}

\def\takeTheRest#1\ofTheString\fi
{\fi \xHyphenate#1\wholeString}

\def\transform#1{\url{#1}\hskip 0pt plus 1pt}
\def\mbf {}

\definecolor{dkgreen}{rgb}{0,0.6,0}
\definecolor{gray}{rgb}{0.5,0.5,0.5}
\definecolor{mauve}{rgb}{0.58,0,0.82}
\definecolor{lightsilver}{rgb}{0.96,0.96,0.98} 
\definecolor{blond}{rgb}{1, 0.98, 0.8}
\lstdefinestyle{R} {
backgroundcolor=\color{blond},
language=Python,
basicstyle=\tiny,
} 
\titleformat{\chapter}[display]
{\normalfont\Huge\bfseries\raggedright}{\chaptertitlename\ 3}
{15pt}{\Huge}
 
\author{Justin London}
\date{December 20, 2024}

\begin{document}

\title{Neural Artistic Style and Color Transfer Using Deep Learning}
\author{\IEEEauthorblockN{Justin London}
\IEEEauthorblockA{\textit{Department of Electrical Engineering and Computer Science} \\
\textit{University of North Dakota}\\
Grand Forks, North Dakota USA \\
justin.london@und.edu}
}
\maketitle

\begin{abstract}
     Neural artistic style transfers and blends the content and style representation of one image with the style of another. This enables artists to create unique innovative visuals and enhances artistic expression in various fields including art, design, and film. Color transfer algorithms are an important in digital image processing by adjusting the color information in a target image based on the colors in the source image.  Color transfer enhances images and videos in film and photography, and can aid in image correction.  We introduce a methodology that combines neural artistic style with color transfer. The method uses the Kullback-Leibler (KL) divergence to quantitatively evaluate color and luminance histogram matching algorithms including Reinhard global color transfer, iteration distribution transfer (IDT), IDT with regrain, Cholesky, and PCA between the original and neural artistic style transferred image using deep learning.  We estimate the color channel kernel densities. Various experiments are performed to evaluate the KL of these algorithms and their color histograms for style to content transfer.  
\end{abstract}
\title{Neural Artistic Style and Color Transfer Using Deep Learning}
\author{Justin London}

\maketitle



\begin{figure}
    \centering
	\includegraphics[width=0.7\columnwidth]{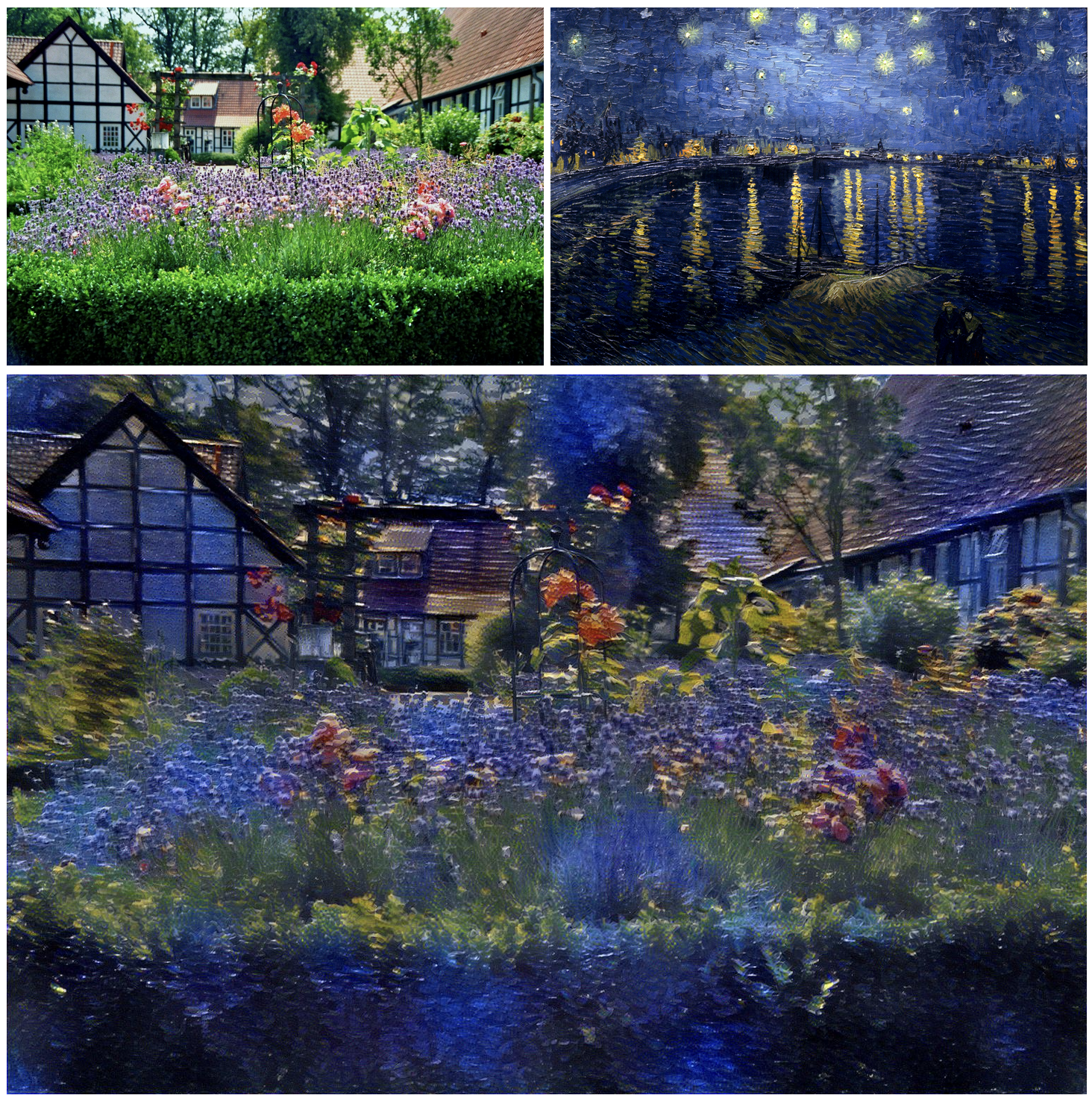} 
	\caption{Illustration of style transfer, but not color.  \cite{Gatys:2015}}
	\label{fig:prop2} 
\end{figure}

\section{Introduction}

    \ \ \ Neural artistic style transfer is a deep learning method, proposed by Gatys, et. al. \cite{Gatys:2015, Gatys2:2016}, that allows the artistic style in a style image ($S$) to be transferred to a content image ($C$).  Neural artistic style transfer has applications in generative AI of digital art and the creation of synthetic artwork from photographs.   Neural style transfer opens up many possibilities in design and content generation.  For instance, the artistic style of Van Gogh can be \enquote{transferred} from an image of a Van Gogh painting such as Starry Nights (the style image) to another target image that contains the content from a content image, including content unrelated to the style image, such as a garden of flowers or image of a herd of horses grazing in the pasture.    The target image generated, after the neural artistic style transfer, will have the same content, but in the style of Van Gogh as well as the style colors used in the style transfer as illustrated in Figure \ref{fig:prop2}.
    
    The key finding of Gatys, et al. is that \enquote{representations of content and style in the convolutional neural network (CNN) are separable.  That is, we can manipulate both representations independently to produce new, perceptually meaningful images \cite{Gatys:2015}.}   CNNs, using pre-trained for image classification can \enquote{encode perceptual and semantic information about images \cite{Narayanan:2017}.}  The input image is encoded in each layer of the CNN by the filter responses to that image.  The shallow layers of a CNN are used to detect lower-level features such as edges and simple textures.  The deeper layers tend to detect higher-level features such as a more complex textures and object classes.  To make the generated image G match the content image C, one should choose a \enquote{middle} activation layer $a^{(l)}$.  The middle layer is neither too shallow nor too deep.   This ensures that the CNN detects both higher-level and lower-level features.  
    
    Let a CNN layer with $N^{l}$ distinct filters generates $N^{l}$ feature maps of size $M^{l}$ where $M_{l}$ is the height times the width of the feature map.  The responses in layer $l$ are stored in a matrix $F^{l} \in \mathbb{R}^{{N_{l}}\times M_{l}}$ where $F^{l}_{ij}$ is the activation of the $i$th filter at position $j$ in layer $l$.   $P^{l}$ and $F^{l}$ are the feature representations of the original image and the generated image, respectively, in layer $l$.  

    The loss function that is optimized by the CNN can be decomposed into three parts: \textit{content loss}, \textit{style loss}, and \textit{variation loss.}   The relative importance of these losses can be determined by a a set of scalar weights that can be fine tuned to get optimal results.  In particular, The performance of the transfer learning can be measured through optimization of the following loss:
\begin{equation}
	\mathcal{L}_{total} = \alpha\mathcal{L}_{content}(\boldsymbol{c},\boldsymbol{x})  + \beta\mathcal{L}_{style}(\boldsymbol{z},\boldsymbol{x}) 
    \label{eq:total} 
\end{equation} 
for content and style weight parameters $\alpha > 0$ and $\beta > 0$, respectively, where the content loss is defined as the squared-error loss between the two feature representations
\begin{equation}
	\mathcal{L}_{content} = \frac{1}{2} \sum_{i=1}^{N_{l}} \sum_{j=1}^{N_{l}} (F^{l}_{ij} - P^{l}_{ij})^{2} 
\end{equation}

    $\boldsymbol{c}$ represents a content image, $\boldsymbol{z}$ represents the (art) style image, and $\boldsymbol{x}$ represents the generated combination image. 
    
	To generate a texture that matches the style of a given input image through style reconstruction, Gatys performs gradient descent from a white noise image to find another image that matches the style representation of the original image.  This is accomplished by minimizing the mean-squared distance between the entries of the Gram matrix from the original image $\boldsymbol{z}$ and the Gram matrix of the generated image $\boldsymbol{x}$.    
    There are two main approaches to texture synthesis: \enquote{non-parametric methods which synthesize a texture by extracting pixels(or patches) from a reference image that are resampled for rendering, and parametric models which optimize reconstructions to match certain statistics computed on filter responses \cite{Berger:2016}.}  
	
	 The style representation \enquote{computes the correlations between the different filter responses where the expectation is taken over the spatial extend of the input image.  These features are given by the Gram matrix $\boldsymbol{G}^{l} \in \mathbb{R}^{{N_{l}} \times N_{l}}$, where $G^{l}_{ij}$ is the inner product between the vectorized feature map $i$ and $j$ in layer $l$ \cite{Gatys:2015}.}    Let $A^{l}$ and $G^{l}$ as the style representations of the original image and generated image at layer $l$. 
\begin{equation}
	G^{l}_{ij} = \sum_{k} F^{l}_{ik}F^{l}_{jk} 
\end{equation} 
	The contribution of that layer to the total loss is
\begin{equation}
	E_{l} = \frac{1}{4N^{2}_{l}M^{2}_{l}} \sum_{i}^{N^{l}} \sum_{j}^{M^{l}} (G^{l}_{ij} - A^{l}_{ij})^{2} 
\end{equation} 
and the total loss is
\begin{equation} 
	\mathcal{L}_{\text{style}}(\boldsymbol{x}, \boldsymbol{z}) = \sum_{l=0}^{L} w_{l}E_{l} 
\end{equation} 
where $w_{l}$ are weighting factors of the contribution of each layer to the total loss.  Gatys, et. al. \cite{Gatys:2015, Gatys2:2016}  use $w_{l} = 0.2$ for matching style representations on convolutional layers $'conv1\_1', 'conv2\_1','conv3\_1', 'conv4\_1$, and $'conv5\_1'$ and $w_{l} = 0$ in all other layers.   

A total variational loss can also be include in the loss function in equation \ref{eq:total}.  This loss is designed to keep the generated image visually coherent and by imposing local spatial continuity between the pixels of the combination image $\mbf{x} = \mbf{f}(x,y)$.  Total variational loss is used to reduce high frequency artifacts and is analogous to an explicit regularization term on the high-frequency components of the combination image. 
\begin{equation} 
	\mathcal{L}_{\text{total\_variation}}(\mbf{f}) = \gamma \sum_{i,j \in \mathcal{N}}\lvert f_{i+1,j} - f_{i,j} \rvert^{\kappa} + 
    \lvert f_{i,j+1} - f_{i,j} \rvert^{\kappa}  
\end{equation} 
where $\mathcal{N}$ is the pixel neighborhood of row pixels $i=1,\dots,N$ and column pixels $j=1,\dots,M$, $\gamma$ is the total variational weight, $\kappa$ is either 1 or 2, $N$ is the number of rows, and $M$ is the number of columns of the combination image $\mbf{f}$. Total variational loss has also been shown to be an effective loss function in unsupervised deep learning for image semantic segmentation.  In this case, the unsupervised loss function is the sum of the absolute values of the gradient of the image in the x and y direction, respectively, using vectorized Sobel operators with respect to $3 \times 3$ spatial offset of pixel $x_{k}$ in neighborhood $\mathcal{N}$ \cite{Javanmardi:2018}.  In other words, the Sobel operator is used on vector of size $9 \times 1$ containing the $3 \times 3$ pixel neighborhood of pixel $x_{k}$. 

    A VGG-16 CNN network can be used to learn the style transfer by minimizing the total loss during gradient descent optimization due to its robust performance in image processing.   The VGG-16 CNN will generate an output image that will have similar textures and colors as the colors are not preserved in the content image.
 
    While the image style transfer method produces high-quality images, it is computationally expensive since each stop of the optimization problem requires a forward and backward pass through the pretrained network \cite{Johnson:2016}.   To overcome this computational burden, Johnson, et. al. (2016) \cite{Johnson:2016} propose the use of perceptual loss functions  for training a feed-forward CNN that solves the optimization problem proposed by Gatys, et. al. in real time.  
    Rather than use \textit{per-pixel} loss functions that depend only on low-level pixel information, the perceptual loss functions depend on on high-level image features extracted from pretrained CNNs.   Thus, their method generates similar quality output in image transformation tasks, but is faster.
 
    The loss of color preservation may lead to a target image that does not posses the intended colors distorting the artistic vision as illustrated in Figure \ref{fig:prop1} where the target image contains the darker colors in the style image, but loses the liveliness and sharp colors in the content image.   However, There are several methods to preserve the vivid and colorful garden and flower colors in the content image as illustrated in Figure \ref{fig:prop2}.  
	\begin{figure}[H]
    \centering \includegraphics[width=0.7\columnwidth]{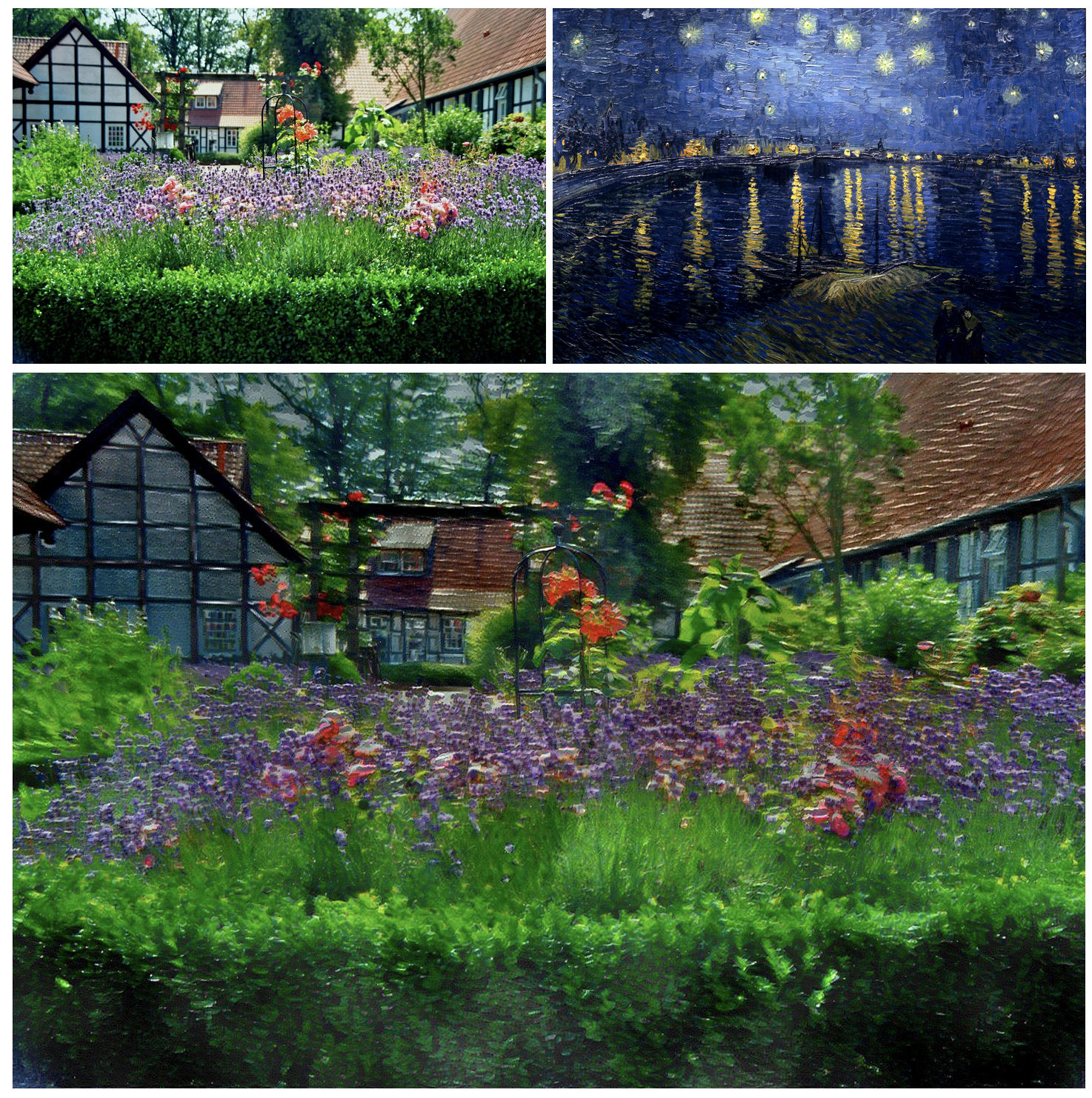} 
	\caption{Illustration of content and style transfer.  \cite{Gatys:2015}}
	\label{fig:prop3} 
    \end{figure} 

    Color transfer has various applications including color balancing and color grading for image and film editing.  Image editing includes image adjustment, image enhancement, image stitching, gamut mapping, image recognition, and image collage \cite{Liu:2022}.  Manually adjusting colors to match them between frames is a delicate task since the change in lighting conditions induces a very complex change of illumination \cite{Pitie:2007}.  The iterative distribution transfer (IDT) algorithm proposed by Piti$\acute{e}$ et. al. automates this process and is a re-coloring method.  

\begin{figure}[H]
    \centering
\includegraphics[width=0.8\columnwidth]{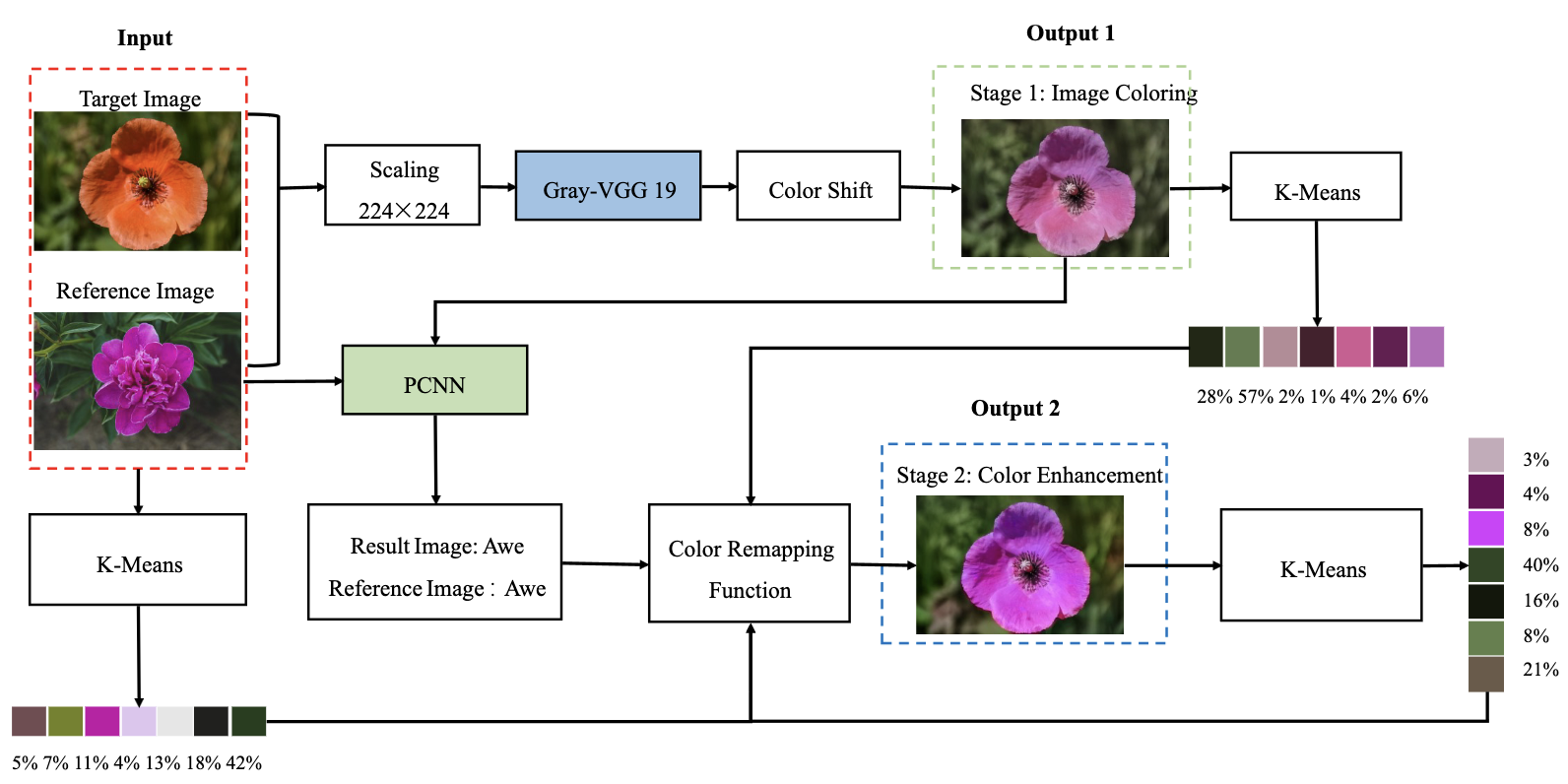} 
	\caption{Two-stage CNN Model for Color Transfer. \cite{Xu:2022}}
	\label{fig:cnntransfer} 
    \end{figure}

    CNNs are powerful albeit computationally expensive method to transfer color and achieve superb results.  Xu and Ding \cite{Xu:2022} introduce a new deep color transfer algorithm based on a two-stage CNN.  The first stage is based on a VGG19 architecture backbone like in Gatys et. al. A reference image-based color transfer (RICT) model is used to extract the features of the reference image and the target image to perform the color transfer.  The second stage is a based on progressive CNN (PCNN) backbone.  They use a \enquote{palette-based emotional color enhancement (PECE) model to adopt the emotional coloring of the generated image by comparing the palette, emotional value and the proportion of each color of the reference image \cite{Xu:2022}.}  An illustration of their model is shown in Figure \ref{fig:cnntransfer}.
    
\section{Color Transfer Algorithms} 

    \ \ \ Color transfer methods can be classified into three categories based on statistical information, geometry, and based on user interaction \cite{Liu:2022}.  Tramsfer methods can either be local or global in nature, meaning the colors are transferred only from certain image regions or from the entire image palette, respectively.
    
    Color transfer can serve two different purposes in film and photography.  The first is to calibrate the colors of two digital cameras for further processing using two or more sample images.  Second, to adjust the colors of two images for perceptual visual compatibility.  For instance, Li et. al. \cite{Li:2014} propose a color transfer algorithm based on color combination that can evoke different emotions.  This algorithm can transfer different emotions through various color schemes.  Unlike other color transfer methods, this method has no reference image, and the color transfer results in a single emotional image \cite{Xie:2020}. 
    
    Xie et. al. propose a transfer algorithm that includes color classification mapping of the target image from the source image using $k$-means algorithm. \cite{Xie:2020}.  Grogan et. \cite{Grogan:2015, Grogan:2019} adopt a method based on shape registration while Wang et. al. \cite{Wang:2017}  propose a gradient-preserving algorithm based on the L0 norm.  These methods are local in nature where colors are transferred based on regions.  
    
    This project will focus on global transfer methods where the image is not segmented and regions are not not clustered based on color and thus the transfer is not based on localized regions.

\subsection{Reinhard Global Color Transfer}

    \ \ \ The Reinhard algorithm matches the mean and variance of the target image to the source image using an affine transformation \cite{Reinhard:2001}.   The transfer of statistics is performed separately for each channel. Since the RGB color space is highly correlated, the transfer is done in another color space $l \alpha \beta$ that has uncorrelated components and better accounts for human-perception of color.  
    
    $\ell \alpha \beta$ is a transform of LMS cone space.  Reinhard et. al. \cite{Reinhard:2001} shows how this can be done in two steps.  First, RBG values are  converted to XYZ tristimulus values and then XYZ are in turn converted to LSM cone space.  Since data in the color space exhibits a lot of skew, one converts the LMS to logarithmic space, i.e. $L = logL, M=logM, and S = logS$.   To decorrelate these axes, Ruderman et. al. \cite{Ruderman:1998} uses PCA which effectively rotates the them resulting in orthogonal principal axes.
        Once the RBG is translated into $l \alpha \beta$ space, one first subtracts the mean:
    \begin{align}
        l^{*} &= \ell - \mu_{l} \\
        \alpha^{*} &= \alpha - \mu_{\alpha} \\
        \beta^{*} &= \beta - \mu_{\beta}
    \end{align}
    Then these points comprising the synthetic image are scaled by their standard deviations:
    \begin{align}
        l^{'} &= \frac{\sigma^{l}_{t}}{\sigma^{l}_{s}}l^{*} \\
        \alpha^{'} &= \frac{\sigma^{\alpha}_{t}}{\sigma^{\alpha}_{s}}\alpha^{*} \\
        \beta^{'} &= \frac{\sigma^{\beta}_{t}}{\sigma^{\beta}_{s}}\beta^{*}
    \end{align}
    After this transformation, instead of adding back the averages previously subtracted, one adds the averages computed for the image.   Finally, one converts the results back to RGB via log LMS, LMS, and XYZ color spaces using equations provided by Reinhard, et. al.  Improvement on this algorithm is to transfer covariance information among the channels by aligning the principal component axes, but is still limited to affine transformations of the color space.  

\subsection{Iteration Distribution Transfer}

    \ \ \ \ Piti$\acute{e}$, et. al. (2007) \cite{Pitie:2007} introduced a method for grading the colors between different images.   The first stage of the method finds a 1-to-1 color mapping that transfers the palette of a target image to the original source image.  This is performed using a non-linear algorithm known as iterative distribution transfer (IDT) that transforms an $N$-dimensional probability density function into another one.   The second stage of the method is to reduce grain artifacts through an efficient post-processing algorithm intended to preserve the gradient field of the original source image.

    Let the RGB vector of pixel $i$ in the source image: $\mathbf{u}_{i} = (R_{i}, G_{i}, B_{i})$
    Assume that the original source image color samples are generated from a continuous color probability density function (pdf) $f$ and that the target color samples are generated from a continuous color pdf $g$.   The mapping problem from the source palette to the target palette is to find a differentiable mapping $t$ that transforms the original color pdf $f$ into a new color pdf that matches the target pdf $g$. In 1-dimension, the distribution transfer problem has a simple solution.

    The differentiable mapping $t$ generates the following constraint which corresponds to a change of variables:
    \begin{equation}
        f(u)\text{d}u = g(v)\text{d}v \ \ \text{with} \ t(u) = v
        \label{eq:diff} 
    \end{equation} 
    Integration of both sides of the equality:
    \begin{equation}    
        \int_{u} f(u)\text{d}u = \int_{v} g(v)\text{d}v
    \end{equation} 
    Let $F$ and $G$ denote the cumulative pdf for $f$ and $g$, respectivey, yielding the mapping $t$ expression:
    \begin{equation}
        \forall u \in \mathbb{R}, \ t(u) = G^{-1}(F(u)) 
        \label{eq:transfer} 
    \end{equation}
    where $G^{-1}(\alpha)$ = inf$\{u \lvert G(u) \geq \alpha\}$    The mapping can be solved using discrete look-up tables.

    Extending the 1-dimensional case to $N$-dimensions is not trivial.  However, via the Radon transform, any $N$-dimensional problem can be uniquely formulated as a series of projections onto 1-dimensional axes, thereby resulting in series of 1-dimensional marginal pdfs.   Thus, manipulations of an $N$-dimensional pdf can be accomplished through manipulations on the series of 1-dimensional marginal pdfs so that they match the corresponding marginals of the target distribution. 

\begin{figure}[H]
    \centering
	\includegraphics[width=0.75\columnwidth]{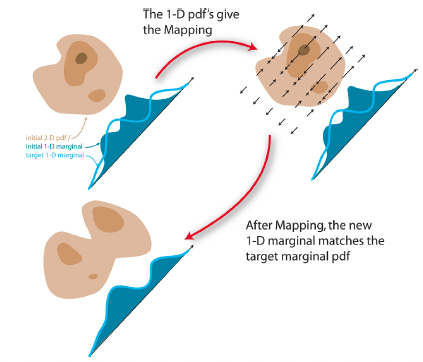} 
	\caption{Data manipulation for 1-dimensional pdf transfer.  \cite{Pitie:2007}}
	\label{fig:transfer3} 
\end{figure}

\begin{figure}[H]
    \centering
	\includegraphics[width=0.9\columnwidth]{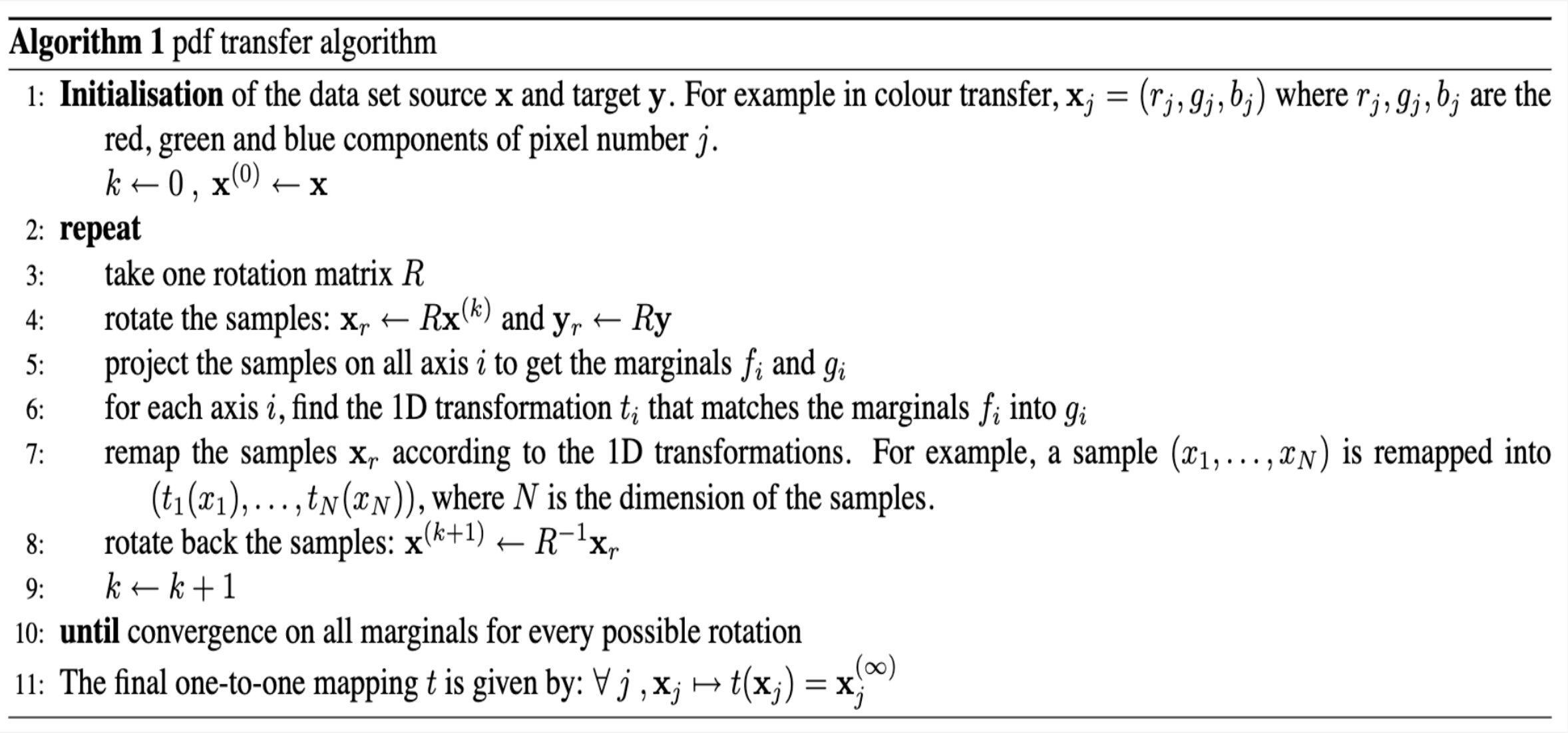} 
	\caption{PDF Transfer Algorithm. \cite{Pitie:2007}}
	\label{fig:pdftransfer} 
\end{figure}

    Let the vector $\mathbf{e} \in \mathbb{R}^{N}$ denote the vector direction of a particular axis.  The projection of both pdfs $f$ and $g$ onto the axis $\mathbf{e}$ results in two 1-dimensional marginal pdfs $f_{e}$ and $g_{e}$.   Using the 1-dimensional pdf transfer mapping in equation \ref{eq:transfer} yields a 1-dimensional mapping $t_{e}$ along this axis:
    \begin{equation}
         \forall u \in \mathbb{R}, \ t_{e}(u) = G_{e}^{-1}(F_{e}(u)) 
    \end{equation} 
    Following Pitie, et. al., for an $N$-dimensional sample $\mathbf{u} = [u_{1},\dots,u_{N}]^{T}$, the projection of hte sample on the axis is given by the dot product $\mathbf{e}^{T}\mathbf{u} = \sum_{i}e_{i}u_{i}$, and the corresponding displacement along the axis is:
    \begin{equation} 
        \mathbf{u} \rightarrow \mathbf{u} + (t_{e}(\mathbf{e}^{T}\mathbf{u}) - \mathbf{e}^{T}\mathbf{u})\mathbf{e} 
    \end{equation} 
    After transformation, the projection $f'_{e}$ of the new distribution $f'$ is now identical to $g_{e}$ as illustrated in Figure \ref{fig:transfer3}.  
    This iterative distribution transfer algorithm is given in Figure \ref{fig:prop1} and 
the pdf transfer algorithm is given in Figure \ref{fig:pdftransfer}. 
\begin{figure}[H]
    \centering
\includegraphics[width=0.6\columnwidth]{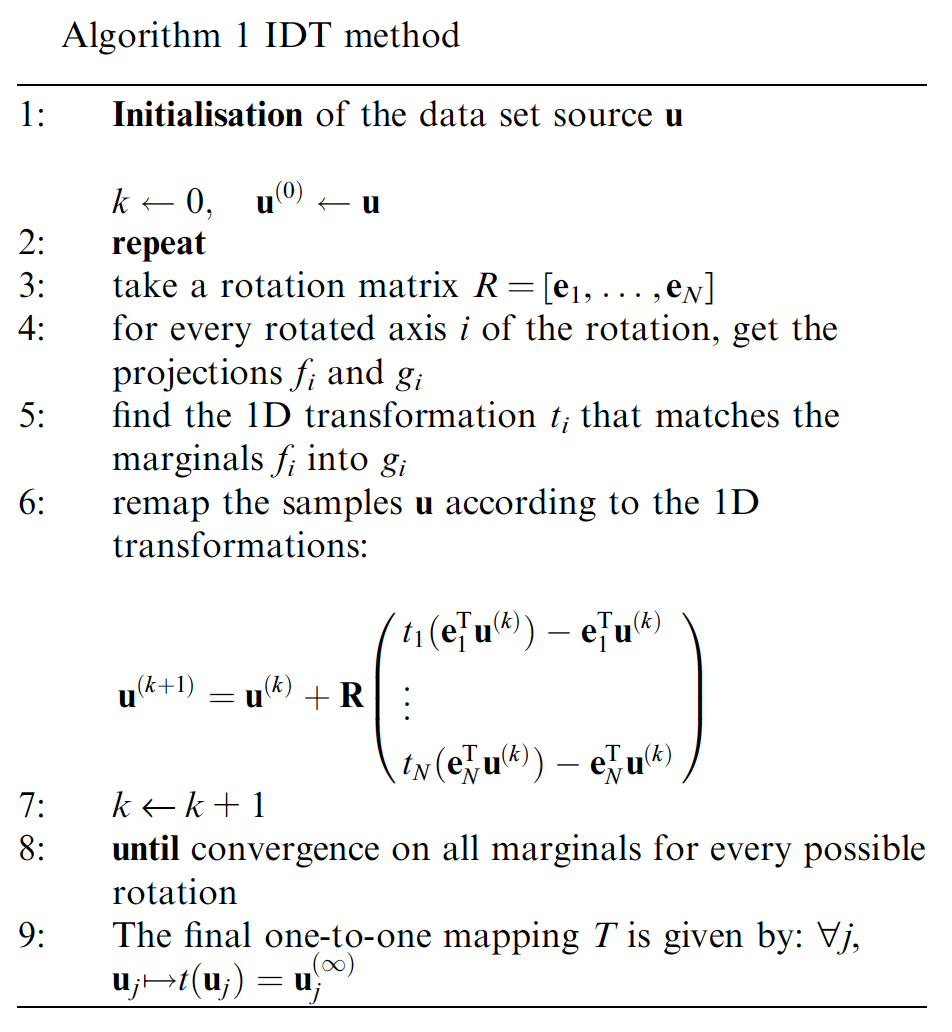} 
	\caption{IDT Algorithm. \cite{Pitie:2007}}
	\label{fig:prop1} 
\end{figure} 
    Piti$\acute{e}$ et. al. show the convergence of the algorithm when the standard when the source and target pdfs are the standard normal distribution:
\begin{align}
    t_{f}: \ \ f \rightarrow \mathcal{N}(\mathbf{0},\mathbf{I}_{N}) \\
    t_{g}: \ \ g \rightarrow \mathcal{N}(\mathbf{0},\mathbf{I}_{N}) \\
    \forall \ \mathbf{u} \in \mathbb{R}^{N}, \ t(\mathbf{u}) = t^{-1}_{g}(t_{f}(\mathbf{u})) 
\end{align} 

    The Kullback-Leibler (KL) divergence, defined as
    \begin{equation}
        D_{KL}(f \lvert \rvert g) = \int_{\mathbf{u}} f(\mathbf{u})\text{ln}\bigg ( \frac{f(\mathbf{u})}{g(\mathbf{u})} \bigg ) d\mathbf{u} 
    \end{equation} 
    can be used as measure to quantity how well the transformed distribution $f^{(k)}$ matches the target pdf $g$. It can handle discrete pdfs by using a summation instead of an integral.  It is a measure of entropy, not a metric for distance.  Kernel density modeling \cite{Silverman:1986} can be used to estimate the underlying pdfs as follows:
    \begin{equation}
        D(f \lvert \rvert g) = \frac{1}{M}\sum_{i=1}^{M} \text{ln} 
        \bigg [\frac{\sum_{j}K\big(\frac{\lvert \rvert \mathbf{u}_{i} - \mathbf{u}_{j} \lvert \rvert}{h_{i}}\big ) }{\sum_{j}K\big(\frac{\lvert \rvert \mathbf{u}_{i} - \mathbf{v}_{j} \lvert \rvert}{h_{i}}\big)} \bigg ] 
        \label{eq:epanech} 
    \end{equation} 
     where $K$ is the Epanechinikov kernel.  The Epanechnikov kernel is the function $K(u) = (3/4)(1-u^{2})$ for $-1 < u < 1$ and 0 otherwise.  To account for the sparseness of samples, is is important to use a variable bandwidth.   Typically, for a sample $\mathbf{u}_{i}$, the bandwidth $h_{i}$ increases with the sparseness of the neighboring samples around $\mathbf{u}_{i}$ \cite{Pitie:2007}.  Comaniciu, et. al. (2001) \cite{Comaniciu:2001} provides an algorithm for a variable bandwidth mean shift to select the scale and propose an adaptive bandwidth $h_{i} = h(\mathbf{x}_{i}) = h_{0} \big [\frac{\lambda}{f(\mathbf{x_{i}}} \big]^{1/2}$ where $h_{0}$ represents a fixed bandwidth, $\lambda$ is a proportionality constant, and $f(\mathbf{x}_{i})$ is the pdf to be estimated.

\begin{figure}[H]
    \centering
    \includegraphics[width=0.7\columnwidth]{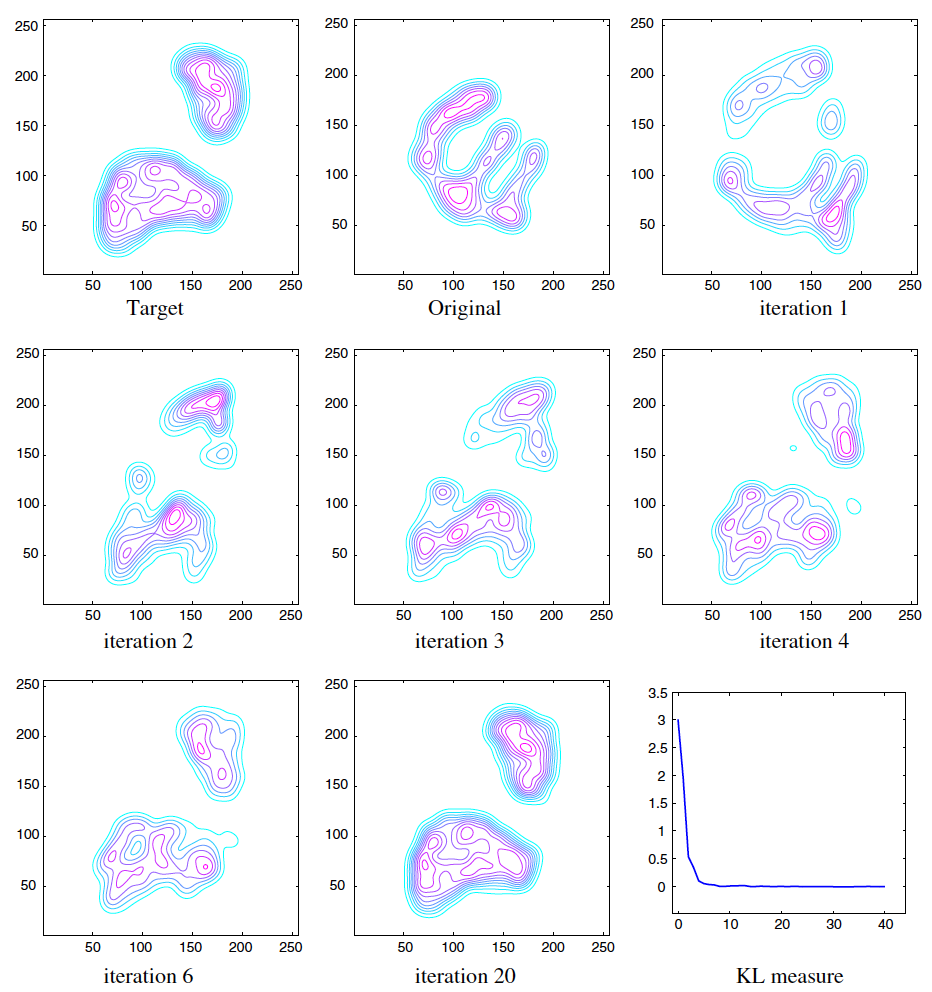} 
	\caption{IDT 2D pdf transfer illustration of convergence.  \cite{Pitie:2007}}
	\label{fig:pdftransfer2} 
\end{figure}

     In the multivariate case, the transfer equation in \ref{eq:diff} can be written as 
\begin{equation} 
    f(\mathbf{u}) = g(t(\mathbf{u}))\lvert \text{det} \ J_{t}(\mathbf{u}) \rvert
\end{equation} 
where $J_{t}(\mathbf{u})$ is the Jacaboian of $t$ taken at $u$.   Linear mappings are in the form $t(\mathbf{u}$) = $\mathbf{Tu}$ + $\mathbf{t}_{0}$ where $\mathbf{T}$ is an $N \times N$ matrix ($N$ = 3 for color).  Thus, the Jacobian is then $J_{t}(\mathbf{u}) = \mathbf{T}$ and the quantity $\lvert \text{det} J_{t}(\mathbf{u}) \rvert = \lvert \text{det} \ \mathbf{T} \rvert$ is a constant.  Thus, $f(\mathbf{u}) \propto g(t(\mathbf{u}))$.  In the case where $f$ and $g$ are multivariate Gaussian distributions (MVG), denoted $\mathcal{N}(\boldsymbol{\mu}_{u}, \mathbf{\Sigma}_{u})$ and $\mathcal{N}(\boldsymbol{\mu}_{v}, \mathbf{\Sigma}_{v})$:
\begin{align}
    f(\mathbf{u}) &\propto \text{exp} \big (\frac{-1}{2}(\mathbf{u} - \boldsymbol{\mu}_{u})^{T} \mathbf{\Sigma}^{-1}_{u}(\mathbf{u} - \boldsymbol{\mu}_{u}) \big) \\
    g(\mathbf{v}) &\propto \text{exp} \big (\frac{-1}{2}(\mathbf{v} - \boldsymbol{\mu}_{v})^{T} \mathbf{\Sigma}^{-1}_{v}(\mathbf{v} - \boldsymbol{\mu}_{v}) \big) 
\end{align}
where $\mathbf{\Sigma}_{u}$ and $\mathbf{\Sigma}_{v}$ are the covariance matrices of $\mathbf{u}$ and $\mathbf{v}$. 
    For the transfer equation to hold, it must be the case 
\begin{multline}
    (t(\mathbf{u}) - \boldsymbol{\mu}_{v})^{T}\mathbf{\Sigma}^{-1}_{v}(t(\mathbf{u}) - \boldsymbol{\mu}_{v}) = \\
    (t(\mathbf{u}) - \boldsymbol{\mu}_{u})^{T}\mathbf{\Sigma}^{-1}_{u}(t(\mathbf{u}) - \boldsymbol{\mu}_{u})
\end{multline}
    Thus, $t$ must satisfy $t(\mathbf{u}) = T(\mathbf{u} - \boldsymbol{\mu}_{u}) + \boldsymbol{\mu}_{v}$ with
$\mathbf{T}^{T}\boldsymbol{\Sigma}^{-1}_{v}\boldsymbol{T} = \boldsymbol{\Sigma}^{-1}_{u}$, or equivalently,
$\mathbf{T}\boldsymbol{\Sigma}^{-1}_{u}\mathbf{T}^{T} = \boldsymbol{\Sigma}_{u}$ \cite{Pitie:2007b}.
    Figure \ref{fig:pdftransfer2} illustrates a 2-D pdf transfer using IDT.  The decrease of the KL divergence demonstrates convergence of the IDT method.

\subsection{IDT with Regrain}

 \ \ \ Color mapping can produce some grain artifacts on the generated image especially if the original image and the target palette suffer from flicker due to extreme brightness variation.  The mapping transformation can also lead to overstretching, which results in an increased level of noise on the mapped original frame.  
 
To reduce grain noise artifacts, Piti$\acute{e}$ et. al. run a post-processing IDT algorithm that forces the noise level to remain the same between the original and target images.  This is accomplished by adjusting the gradient field of the generated target image so that it matches the gradient field of the original image.  Piti$\acute{e}$ et al. note manipulation of the image gradient can be achieved using a variation approach.  However, since re-coloring implies changes in the contrast levels, the new gradient field should only loosely match the original gradient field \cite{Pitie:2007}. 

    Denote $I(x,y)$ as the three-dimensional original color image.  Let $t(I)$ be the color transformation. The problem is to find a modified image $J$ of the mapped image $t(I)$ that minimizes the entire image range $\Omega$:
    \begin{equation}
        \underset{J}{\text{min}} \int \int_{\Omega} \phi \cdot \lvert \rvert \nabla J - \nabla I \lvert \rvert^{2} + \psi \cdot \lvert \rvert J - t(I) \lvert \rvert^{2}\text{d}x\text{d}y
        \label{eq:lagrange} 
    \end{equation}
    with Neumann boundaries condition $\nabla J \lvert_{\partial \Omega} = \nabla I \lvert_{\partial \Omega}$ so that the gradient of $J$ matches the gradient of $I$ at the image border $\partial \Omega$.  The term $\lvert \rvert \nabla J - \nabla I \lvert \rvert^{2}$ forces the image gradient to be preserved while the term $\lvert \rvert J - t(I) \lvert \rvert^{2}$ ensures that the colors remain close to the target image thereby protecting the contrast changes.

    The weight fields $\phi(x,y)$ and $\psi(x,y)$ impact the relative importance of each term. While many choices are possible for the weights, Piti$\acute{e}$ selects the first weight $\phi$ emphasize that only flat areas have to remain flat but that the gradient can change at object borders:
    \begin{equation}
        \phi = \frac{30}{1 + 10 \lvert \rvert \nabla I \lvert \rvert}
    \end{equation}  
    The second weight field $\psi(x,y)$ account for the possible stretching of the transformation $t$.  When $\nabla t$ is large, the grain becomes more visible:
    \begin{equation}
        \psi = 
        \begin{cases}
            & 2/(1 + \lvert \rvert (\nabla t)(I) \lvert \rvert) \ \ \text{if} \ \ \lvert \rvert \nabla I \lvert \rvert > 5 \\
            & \lvert \rvert \nabla I \lvert \rvert / 5 \ \ \ \ \ \ \ \ \ \text{if} \ \ \lvert \rvert \nabla I \lvert \rvert \leq 5
        \end{cases} 
    \end{equation}  
    where $(\nabla t)(I)$ is the gradient of $t$ for color $I$ and refers to color stretching.  The case $\lvert \rvert \nabla I \lvert \rvert \leq 5$ \ is necessary to reinforce that flat areas remain flat.  

    Piti$\acute{e}$ et al. shows that the minimization of equation \ref{eq:lagrange} can be solved using the variational principle which states that the integral must satisfy the Euler-Lagrange equation:
    \begin{equation}
        \frac{\partial F}{\partial J} - \frac{\text{d} \ \partial F}{\text{dx} \ \partial  J_{x}} - \frac{\text{d} \ \partial F}{\text{dy} \ \partial J_{y}} = 0
    \end{equation}
    where $F(J,\nabla J) = \lvert \rvert \nabla J - \nabla I \lvert \rvert^{2} + \psi \cdot \lvert \rvert J - t(I) \lvert \rvert^{2}$,
    from which the following elliptical partial differential equation is derived:
    \begin{equation}
        \phi \cdot J - \text{div}(\phi \cdot \nabla J) = \phi \cdot t(I) - \text{div}(\phi \cdot \nabla I) 
    \end{equation}  
    Numerical solutions to solve this 2D system (functions of x and y) can be solved by standard iterative methods such as SOR and Gauss-Sidel with multigrid approach.

\begin{figure}[H]
    \centering
\includegraphics[width=0.8\columnwidth]{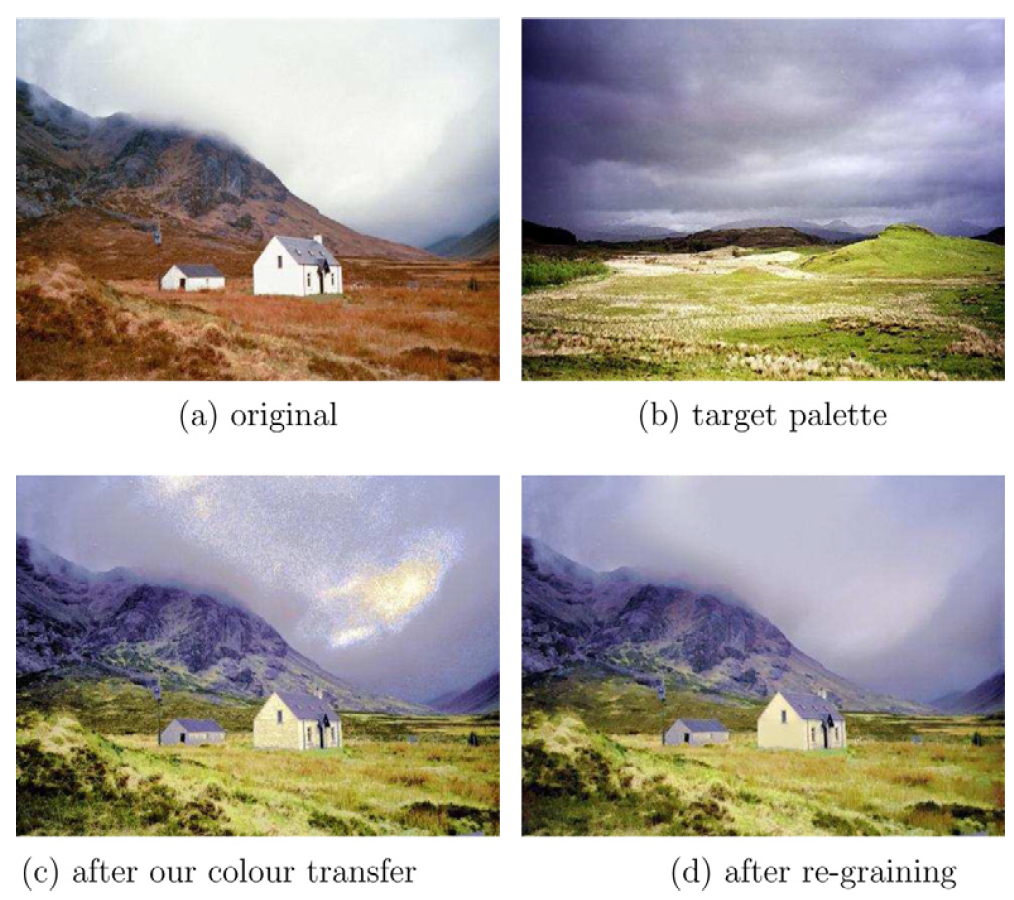} 
	\caption{IDT with regraining. \cite{Pitie:2007}}
	\label{fig:regrain} 
\end{figure}

    Figure \ref{fig:regrain}(c) shows artifact grain after the color transfer from the targette palette in (b) to the original image (a).  After re-graining to reduce the noise, the details in the original image are preserved, while the spurious graininess in the sky is removed and washed out. 
    
\begin{figure}[H]
    \centering
    \begin{subfigure}{0.4\textwidth}
    \includegraphics[width=\textwidth]{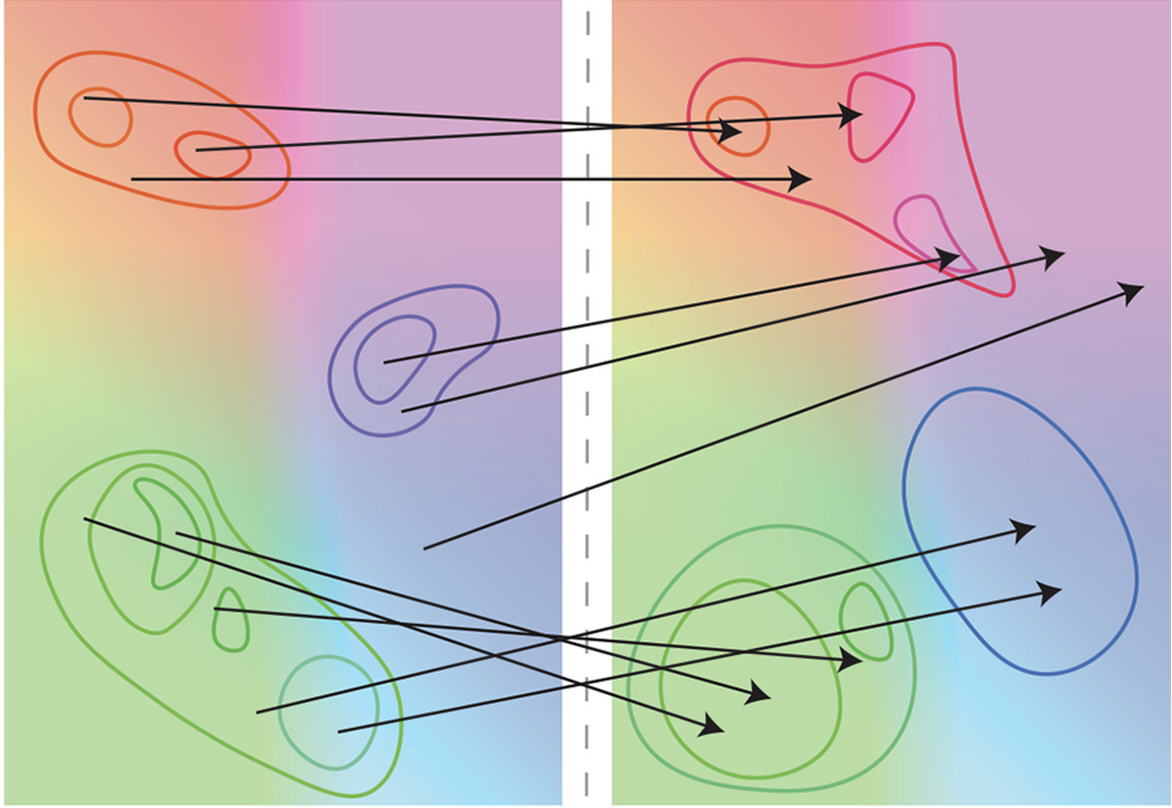} 
	\subcaption{OT Map For Continuous Dist.}
    \end{subfigure} 
    \begin{subfigure}{0.4\textwidth}
    \includegraphics[width=\textwidth]{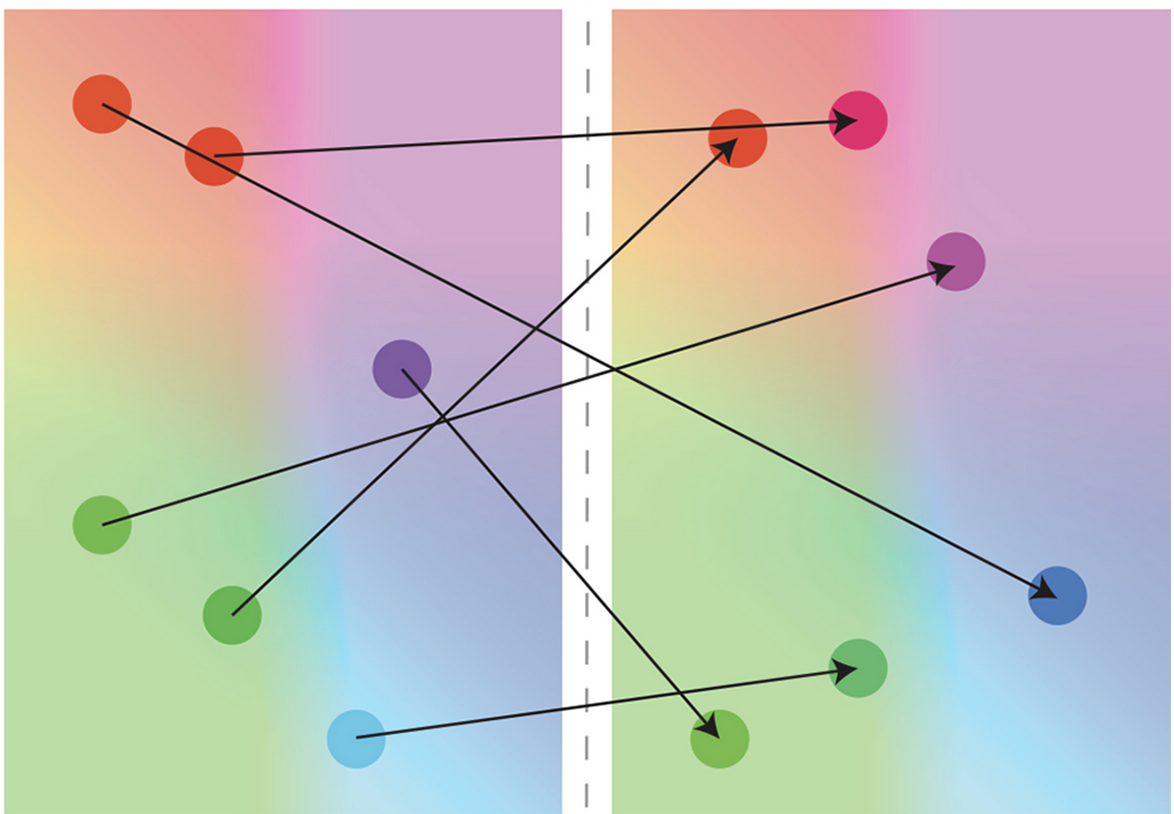} 
    \subcaption{OT Map for Discrete Dist.} 
    \end{subfigure} 
    \caption{\cite{Pitie:2020}} 
	\label{fig:otmap} 
\end{figure}

\subsection{Monge-Kantorovitch Optimal Transport Algorithm}

    \ \ \ One of the drawbacks with affine transformations is that the resulting image could have the color proportions as expected, but locally the colors could be swapped.   For instance, it is not guaranteed that a transfer will not map black pixels to white pixels and vice versa.   To overcome this problem, one can constrain the transfer problem and looking for a mapping that minimizes its displacement cost:
    \begin{equation}
        I[t] = \int_{u} \lvert \rvert t(\mathbf{u}) - \mathbf{u} \lvert \rvert^{2} f(u)\text{d}u
    \end{equation} 
    Finding the solution that minimizes the displacement mapping is Monge-Kantrovich (MK) optimal transportation problem.   The MK solution always exists for continuous pdfs and is \textit{unique}.  Furthermore, since it turns out that the MK solution is the gradient of a convex function, 
    \begin{equation}
        t = \nabla \phi \ \ \text{where} \ \ \phi: \mathbb{R}^{N} \rightarrow \mathbb{R} 
    \end{equation} 
    it is equivalent to monotonicity for 1-dimensional functions in $\mathbb{R}$ \cite{Pitie:2007b}.  This implies that the brightest regions of an image still remain the brightest regions after mapping.  Moreover, the MP map of the gradient of the convex function like $c(\mathbf{u},\mathbf{v}) = \lvert \rvert \mathbf{u} - \mathbf{v} \lvert \rvert^{2}$ means that the optimal color transformation contains no rotation or inversion \cite{Pitie:2020} as illustrated in Figure \ref{fig:grad}:
\begin{figure}[H] 
    \centering
	\includegraphics[width=0.8\columnwidth]{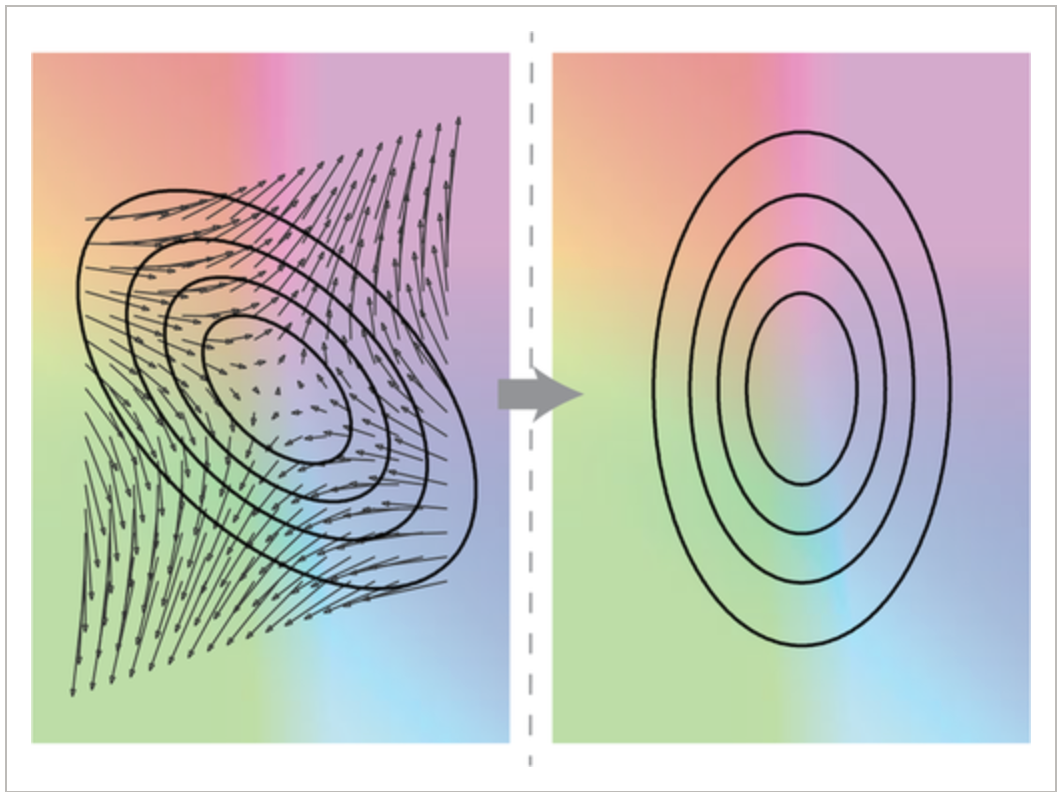} 
	\caption{Source: \cite{Pitie:2020}}
	\label{fig:grad} 
\end{figure} 
    The MK optimal transportation formulation relaxes the 1-to-1 mapping constraint.  Instead of using a mapping $\mathbf{u} \rightarrow t(\mathbf{u})$, one estimates a transportation plan $\pi(u,v)$, a joint pdf of correspondence between $u$ and $v$, that  indicates the proportion of pixels with color $u$ are mapped to a particular color $v$.  The optimal transport map for continuous distributions and discrete distributions is shown in Figure \ref{fig:otmap}.
    
    The OT distance is also known as Wasserstein distance in statistics and earth mover's distance (EMD) in computer vision:
    \begin{equation}
        W^{p}(U,V) = \underset{\pi}{\text{min}}\int_{u,v} \pi(u,v) \lvert \rvert u - v \lvert \rvert^{p} \text{d}u \text{d}v.
    \end{equation} 
    Piti$\acute{e}$ \cite{Pitie:2020}  note that the cost of computing the Kantorovitch OT solution is still very expensive as its complexity grows $O(n^3\text{log}(n))$ as with the number of points $n$. The discretised problem can be numerically solved by linear programming using the simplex algorithm. Several specialised algorithms for solving the transport problem also exist, notably the northwest corner method and the Vogel's approximation. Recently, in a landmark paper, Cuturi showed that a further entropic relaxation of OT could lead to a much more efficient implementation using Sinkhorn's algorithm. 
    For discrete distributions,, the 1-to-1 constraint can pose problems since a single color may need to be mapped into multiple colors. However, with MK mapping, permutations of an assignment such as in Figure \ref{fig:otmap} such as do not change the effect on the transformed distribution. Each permutation results in the same distribution.
    
\subsection{Color Histogram Matching} 

\ \ \ First, color histogram matching involves trying to match the color histograms in the target image to those in the content image for each RGB channel.    There are various color (matching) transformation algorithms.    A linear color transfer is simple and effective for color style transfer.
    
    Let $\mathbf{x}_{i} = (R,G,B)'$ be a pixel of an image.  Each pixel is transformed as:
    \begin{equation}
            \mathbf{x}_{S'}  \leftarrow \mathbf{A}\mathbf{x}_{S} + \mathbf{b} 
    \end{equation} 
    where $\mathbf{A}$ is a 3 $\times$ 3 and $\mathbf{b}$ is a 3 $\times$ 1 vector.
        This transformation is selected so that the mean and covariance of the RGB values in the new style image S' match those of the content image C.    Denote $\mu_{S}$ and $\mu_{C}$ be the mean colors of the style and content images and 
        $\mathbf{\Sigma}_{S}$ and $\mathbf{\Sigma}_{C}$ be the pixel covariances. 

        The mean and covariance of the color pixels are given by $\mu = \sum_{i} \textbf{x}_{i}/N$ and $\mathbf{\Sigma} = \sum_{i} (\textbf{x}_{i} - \mu)(\mathbf{x}_{i} - \mu)'/N$.  $\mathbf{A}$ and $\mathbf{b}$ are chosen to satisfy $\mu_{S'} = \mu_{C}$ and $\mathbf{\Sigma}_{S'} = \mathbf{\Sigma}_{C}$.  These are 
        determined by the constraints:
        \begin{align}
            \mathbf{b} &= \mu_{C} - \mathbf{A}\mu_{S} \\
            \mathbf{A\Sigma_{S}A'} &= \mathbf{\Sigma}_{C}
        \end{align} 

    A family of solutions for $\mathbf{A}$ satisfies these constraints.   Gatys, et. al. \cite{Gatys:2015} considers two variations.   The first uses the Cholesky decomposition:
    \begin{equation}
        \mathbf{A}_{\text{chol}} = \mathbf{L}_{C}\mathbf{L}^{-1}_{S}
    \end{equation}  
    where $\mathbf{\Sigma} = \mathbf{LL'}$ is the Cholesky decomposition of $\mathbf{\Sigma}$. 

    The second variation uses an eigenvalue decomposition of the covariance matrix $\mathbf{\Sigma} = \mathbf{U\Lambda U'}.$  Define the matrix square-root as $\mathbf{\Sigma}^{1/2} = \mathbf{U\Lambda^{1/2}U'}$ where $\mathbf{U}$ is the matrix of eigenvectors and $\mathbf{\Lambda}$ is the diagonal matrix of eigenvalues.  Then the transformation is given by:
    \begin{equation}    
        \mathbf{A} = \mathbf{\Sigma}^{1/2}_{C}\mathbf{\Sigma}^{-1/2}_{S}
    \end{equation} 
    where $\mathbf{\Sigma}^{1/2}_{C}$ is the square-root of the content matrix and  $\mathbf{\Sigma}^{-1/2}_{S}$ is the reciprocal of the square-root of the style  matrix.  This second variation is used in the 3D color matching formulation explored in Image Analogies, discussed in the Appendix.  Image analogies is an alternative method to transfer style to a generated target image using approximating nearest neighbors (ANNs) instead of instead of deep learning.

\subsection{Histogram Equalization} 

    \ \ \ A more precise algorithm involves histogram matching and equalization of the color channels by computing the cumulative distribution for each channel histogram and normalizing it.  Histogram equalization adjusts image intensities to enhance contrast.
    
    For each channel of an image $\mathbf{x} \in \mathbb{R}^{2}$ (with dimensions $M \times N$) one computes the histogram of the pixel intensity values using $\code{imhist}$  Let the number of intensity pixel value of $i$ be $n_{i}$ and the total number of pixels in the image $\bar{N} = M\times N$.  Then, the probability of an occurrence of a pixel of intensity $i, 0 \leq i < L=256$, is $p_{x}(i) = p(x = i) = \frac{n_{i}}{\bar{N}}$.  
    
    Thus, $p$ denotes the normalized histogram of $\mathbf{x}$ with a bin for each possible intensity.   The cumulative distribution function, which is the image's accumulated normalized histogram, is then computed as:
    \begin{equation}
        \text{cdf}_{x}(i) = \sum^{i}_{j=0}p_{x}(x=j) 
    \end{equation} 
    To transform the image to the form $\mathbf{y} = T(\mathbf{x}$ with a flat histogram, one needs a linearized CFD across all intensity values:
        \begin{equation}
            \text{cdf}_{y}(i) = (i+1)K \ \ \ \text{for} \ \ 0 \leq i \leq L-1,
        \end{equation} 
    for some constant $K$.   Thus, we need $y = T(v) = \text{cdf}_{x}(v)$ for $v \in [0,L-1].$   $T$ maps the color levels into the range [0,1] since we are using a normalized histogram of $\mathbf{x}$.   The histogram equalized image $\mathbf{y}$ with pixel intensities $i$ of $\mathbf{x}$ is given by:
    \begin{equation}
        y = T(v) = \text{floor}((L-1)\sum_{n=0}^{v} p_{n}) 
    \end{equation} 
    The general histogram equalization formula is:
    \begin{equation}
        h(v) = \text{round}\bigg ( \frac{\text{cdf}(v) - \text{cdf}_{min}}{(M \times N) - \text{cdf}_{min}} \times (L-1) \bigg )          
    \end{equation}

    The transformation can generate significant changes to the target image's color balance since the relative distributions of the color channels change as a result of applying the algorithm.  However, if the image is first converted to another color space, such as Lab, HSL/HSV color space, then the algorithm can be applied to the luminance or value channel wihtout resulting in changes to the hue and saturation of the image \cite{Naik:2003}. 
    

\subsection{Luminance Only Transfer}

    \ \ \ Second, is a luminance-only style transfer.   This is motivated by the \enquote{observation that visual perception is far more sensitive to changes in luminance than in color.}\footnote{\cite{Gatys:2015},p.}     Following Gatys, the luminance channels $L_{S}$ and $L_{C}$ are first extracted from the style and content images.  The neural style transfer algorithm is applied to these images to produce an output luminance image $L_{T}$.   
 
    Gatys, et. al. uses the YIQ color space where the color information of the content image are represented by the $I$ and $Q$ channels;  these are combined with $L_{T}$ to produce the final color output such as those illustrated in Figure \ref{fig:prop1}(e).
 \begin{figure}[H]
    \centering
	\includegraphics[width=0.9\columnwidth]{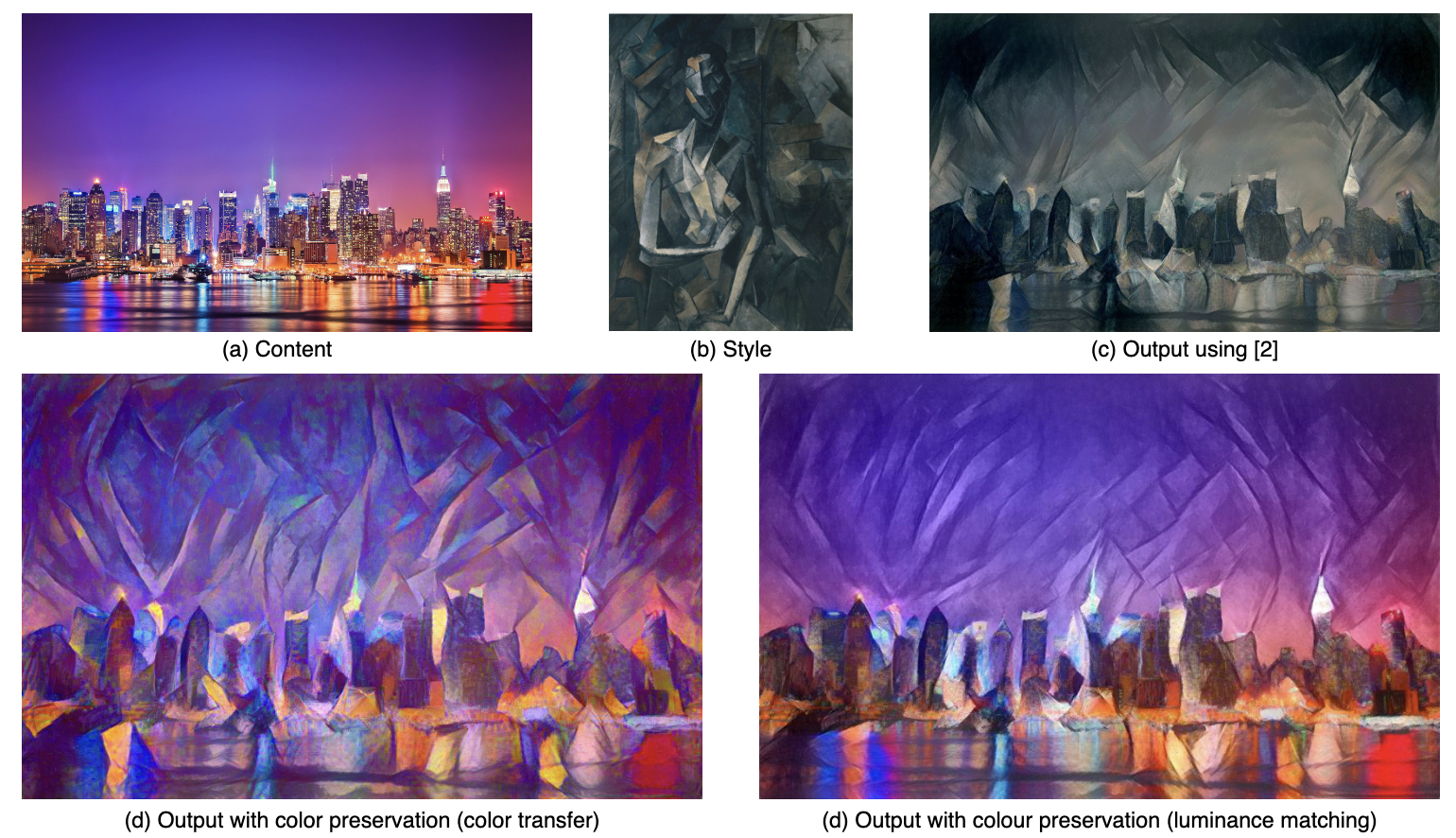} 
	\caption{Illustration of content and style transfer.  Source: \cite{Gatys:2015}}
	\label{fig:prop1} 
\end{figure} 

	One can match the histogram of the style luminance channel $L_{S}$ to that of the content image $L_{C}$ before transferring style if there is a substantial mismatch between their luminance histograms.  Following Gatys, et. al., a linear map can then be used to match the second-order statistics of the content image.   Let $\mu_{S}$ and $\mu_{C}$ be the mean luminances of the style and content images, respectively, and $\sigma_{S}$ and $\sigma_{C}$ be their standard deviations.  Then each luminance pixel in the style image is updated as:
\begin{equation}
	L_{S} = \frac{\sigma_{C}}{\sigma_{S}}(L_{S} - \mu_{S}) + \mu_{C}  
\end{equation}  
    As with IDT, to remove skew, one should convert RGB images to logarithmic (Lab) color space prior to performing the luminance transfer and then covert back to RGB color from Lab space afterwards.  However, in this project, a Python script was used and called within Matlab using the $\code{pyrunfile}$ Matlab function since the Matlab $\code{rbg2lab}$ function does not generate the correct results as the $\code{cv2}$ rgb2lab version in Python.

\section{Methodology} 

    \ \ \ Implementation and adaption of neural artistic transfer using the VGG-19 and the various color transfer algorithms were performed in Matlab \cite{Matlab:2020}.  Due to the complexity of various algorithms, the IDT, regrain, and MLK were adapted and used from Pitie.\footnote{See \url{https://github.com/frcs/colour-transfer}} 
    
    Initially 2500 training iterations were used, but in later experiments it was reduced to 
    1500 iterations because it was observed that no improvement in the training losses or discernible image changes from the style transfer appeared after 1500.  Using less iterations also saved considerable computational time due to the high computational expense required for neural training.  1 iteration takes approximately 3.75 to 4 seconds to complete which means 1500 iterations takes roughly 87.5 to 100 minutes or approximately 1.5 hours to complete.  The learning rate was set to 2.
    
\begin{figure}[H]
    \centering
	\includegraphics[width=0.9\columnwidth]{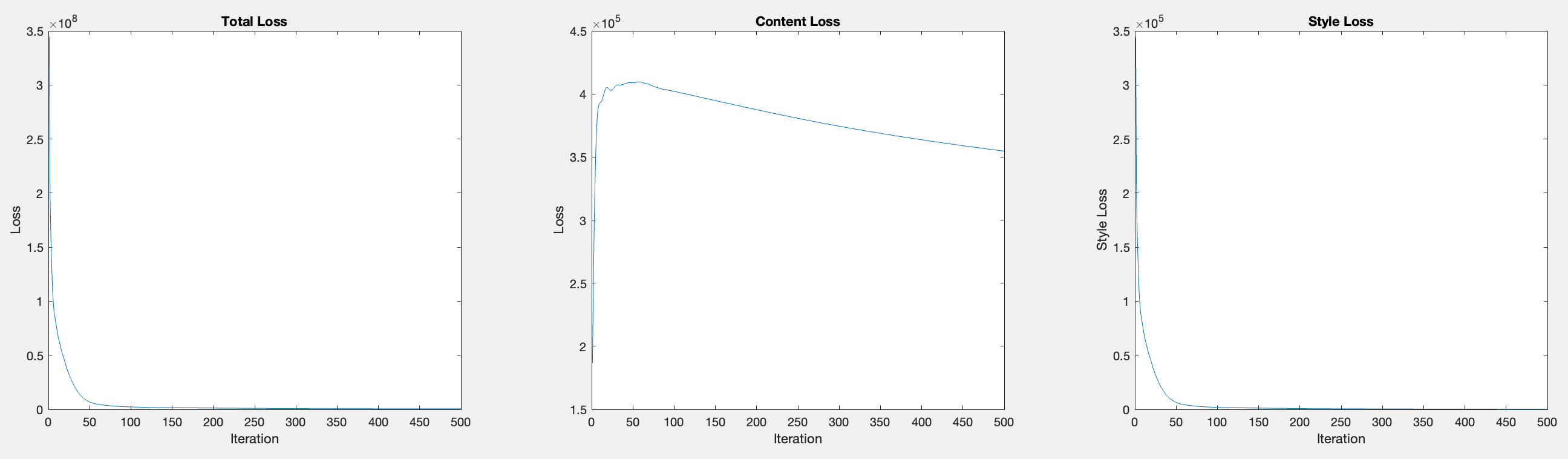} 
	\caption{Training Losses}
	\label{fig:losses} 
\end{figure}

    Figure \ref{fig:losses} shows the total, content, and style training losses  iterations for 500 iterations. With the exception of the content loss, the total loss and style loss converge near zero.  

     The color channel histogram for the content source image was generated.  After the color transform, the images were filtered for sharpening using Matlab's \code{imsharpen}. The color channel histograms for each of the generated images were generated for each of the color transfer algorithms.   
     Kernel density estimation using Matlab's \code{kdensity} function was used to convert the color histograms into probability density functions for each channel.   The Kullback-Leibler (KL) divergence of the kernel pdf density estimate of each color channel with that of the kernel pdf density       
    
    Several experiments were conducted using content images and artistic style images.  Experiment 1 was to transfer the style and color from Monet's \textit{House of Parliament} to a generated target image using a content image of a bridge.  
    Experiment 2 was to transfer the color from this generated target image in Experiment 1 back to that of the content image (target) of the bridge.
    
    Experiment 3 was to transfer the style and color from Van Gough's Starry Night painting to a generated target image using the content image of a lighthouse with a completely different contrast and colors from the style image.   Experiment 4 was to transfer the color from the generated style image (source) in Experiment 3 to that of the content image (target) of the lighthouse.

    The purpose of these experiments is to compare the perceptual quality of the generated images with color and style transfers, their color histograms, kernel density estimates, and ultimately, the KL divergence between the estimated pdfs of generated images and the content images.   

\section{Experimental Results} 
    
\subsection{Experiment 1} 
\begin{figure}[H]
    \centering
    \begin{subfigure}{0.4\textwidth}
    \includegraphics[width=\textwidth]{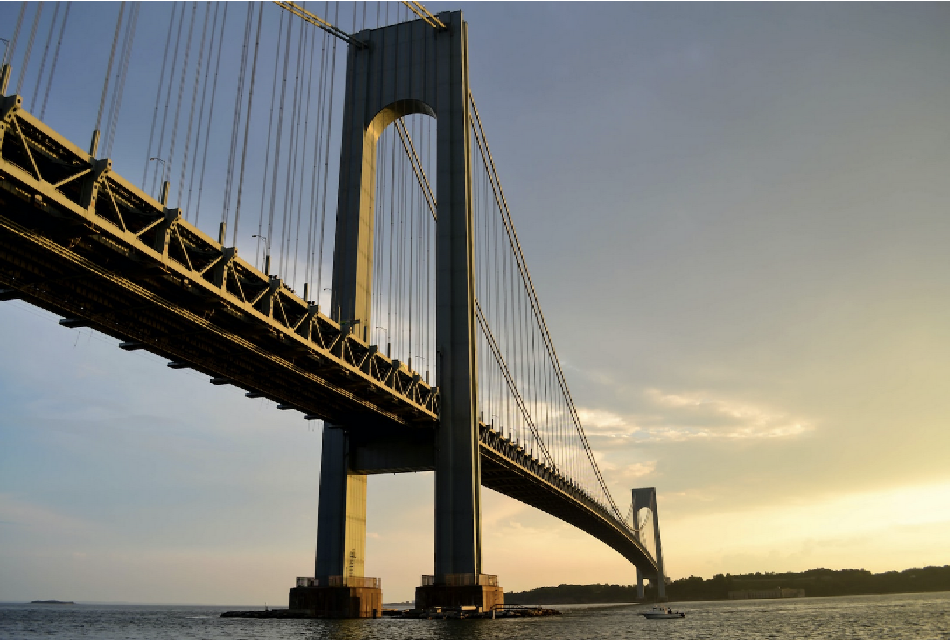} 
	\subcaption{Content Image}
    \end{subfigure} 
    \begin{subfigure}{0.4\textwidth}
    \includegraphics[width=\textwidth]{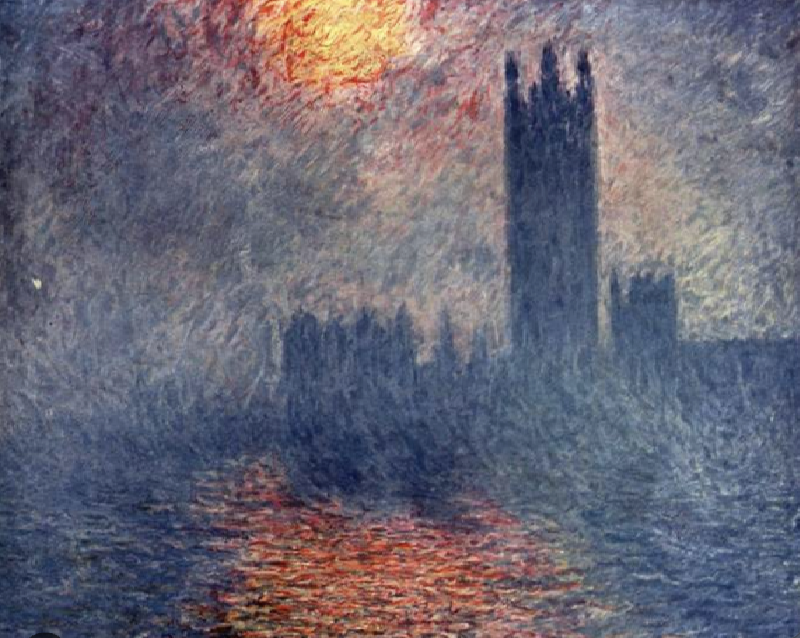} 
    \subcaption{Van Gogh Style Image} 
    \end{subfigure} 
    \caption{} 
	\label{fig:style10} 
\end{figure}

\begin{figure}[H]
    \centering
	\includegraphics[width=0.9\columnwidth]{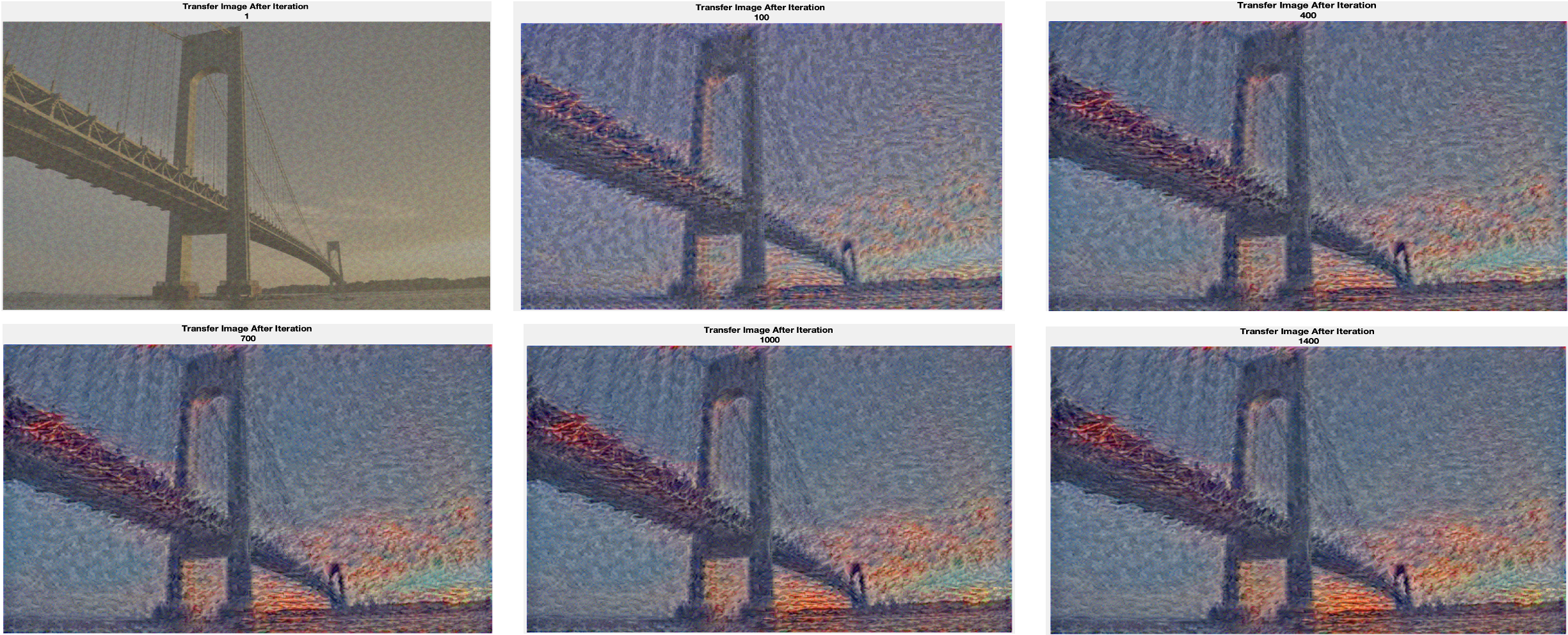} 
	\caption{Color transfer to Generated Monet's Style Image.}  
	\label{fig:bridges}
\end{figure}

\begin{figure}[H]
    \centering
	\includegraphics[width=0.9\columnwidth]{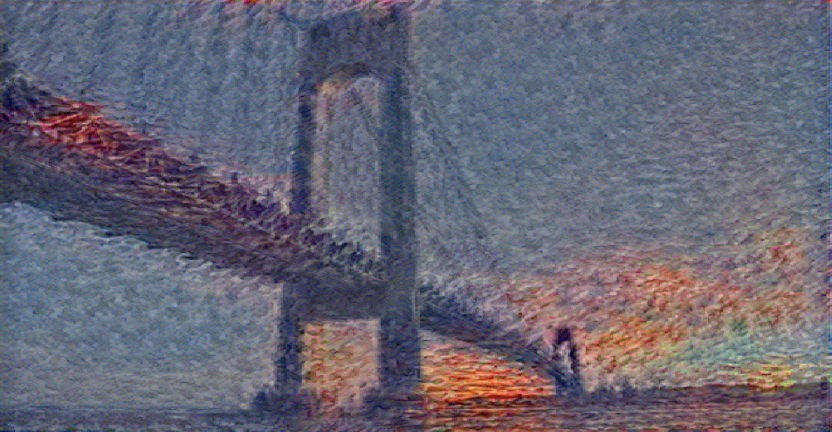} 
	\caption{Final generated style image}
	\label{fig:final} 
\end{figure}

\begin{figure}[H]
    \centering
	\includegraphics[width=0.9\columnwidth]{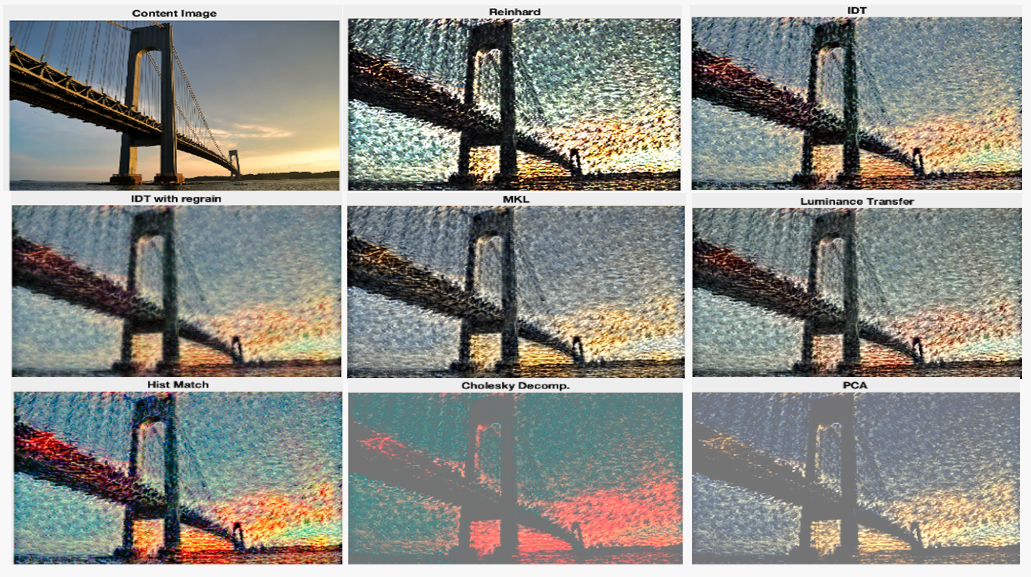} 
	\caption{Color Transfer Generated Images}
	\label{fig:ct} 
\end{figure}

     \ \ \ The content image (bridge) and style image (Monet's \textit{House of Parliament}, London) are shown in Figure \ref{fig:style10}(a) and Figure \ref{fig:style10}(b), respectively. The content image is $950 \times 640 \times 3$ and the style image is $436 \times 346 \times 3$.  Both images have alpha channels.
    Figure \ref{fig:bridges} shows the generated image at different iterations (1,100,300,100,1500,2000) in the neural training.  As shown the high-level (coarse) details are transferred in the early iterations and the more low-level (fine) details are transferred in the higher iterations.   
    The final generated style transfer image is shown in Figure \ref{fig:final} 
    This image is then input into the different color transfer algorithms.  The color transfer generated images are shown in Figure \ref{fig:ct} 
    The top left is the original  content image of the bridge.  The top middle show the image with the color transfer using the Reinhard algorithm.  The top right image shows the color transfer using IDT.   The middle row left images shows color transfer using IDT with regrain, middle center with MKL, and middle right with luminance transfer.  The bottom row left image has histogram matching, bottom middle has Cholesky decomposition, and the bottom right has PCA. 

    As shown, from a perceptual-visual quality, the histogram (equalization) matching appears to provide highest contrast given that the color channels are matched to those in the original image.  The illumination on the bridge and sunset colors are perceptually sharper and brigher than in the other generated images.  
    
    While Reinhard, IDT, IDT with regrain, MKL, and luminance algorithms do not have as sharp a contrast as the histogram matching, they are much better than Cholesky and PCA which have loss in both contrast, brightness, and are very desaturated (the color appears more washed-out or pale).  Moreover, the bluish-grey sky and vanilla sunset colors in the original image transfers in the IDT, IDT with regrain, MKL, and luminance transfer better match the saturation and hue in the content image that the other algorithms.     

    The color histograms for each of the color transfer algorithms is shown in Figure \ref{fig:bridgehist}:
 \begin{figure}[H]
    \centering
	\includegraphics[width=0.9\columnwidth]{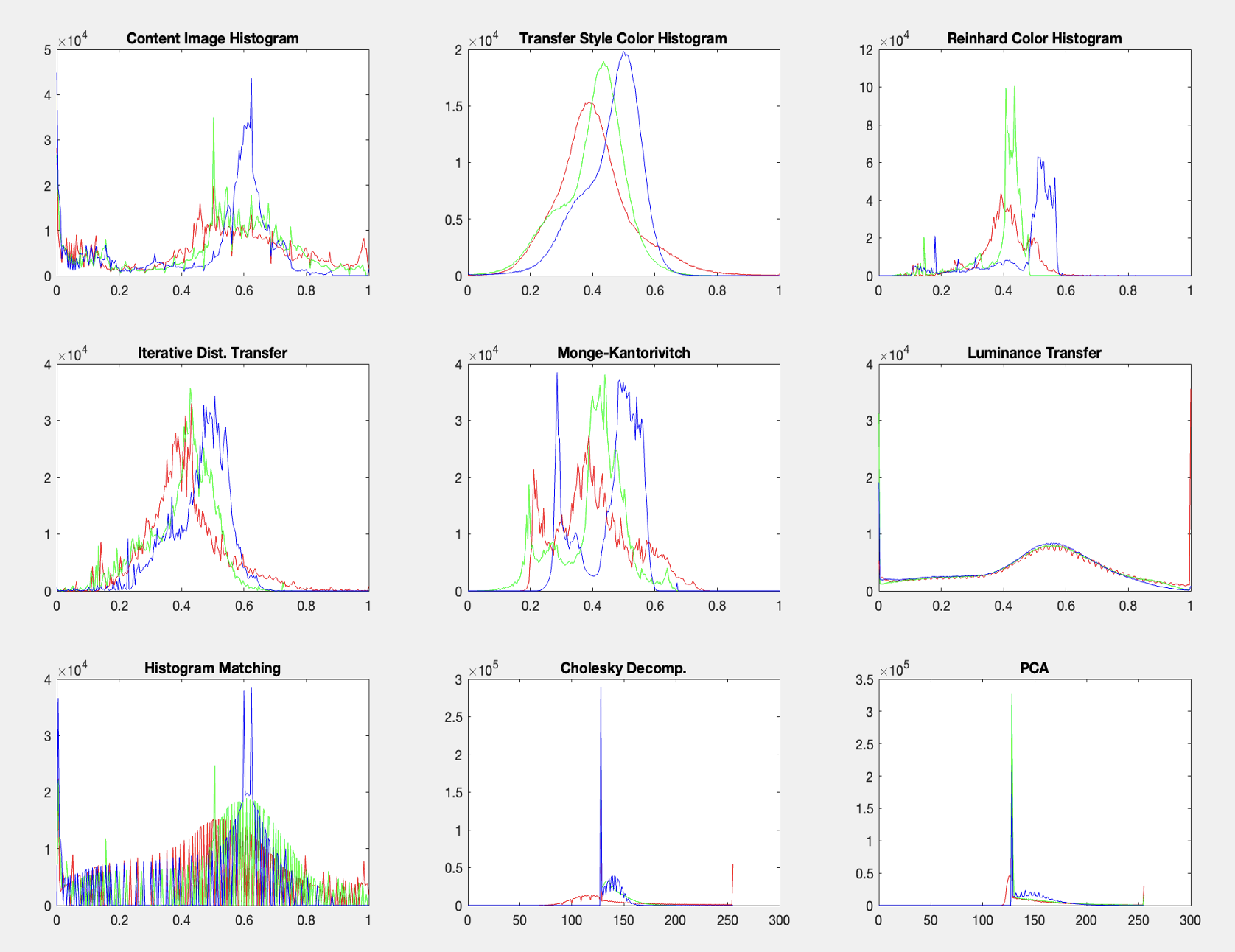} 
	\caption{Color Histograms}
	\label{fig:bridgehist} 
\end{figure}      
    Ideally, we want the color transfer algorithm to have their color histograms match those of the content image (top left).  The Histogram Matching (lower left) matches the content color histograms the closest of all the algorithms followed by the IDT.  This is expected because the histogram matching is actually matching the cumulative distribution function (cdf) of the target image to that of the content source image.  
    
    The Reinhard algorithm exhibits histogram shapes that have peaks like the content image histograms, but they  are not as well aligned with their position, locations, and the dispersion in the tails as the histogram matching algorithm.
    With luminance transfer, all the color channels follow a similar smooth bell-shaped curved shape and exhibit dispersion across the intensity values.  Though luminance transfer does not have the sharp peaks and variability of the content image histograms, it is highlighting the luminance channel, thus extracting and transferring the luminance channel from the content image.   
    
\begin{figure}[H]
    \centering
    \includegraphics[width=0.9\columnwidth]{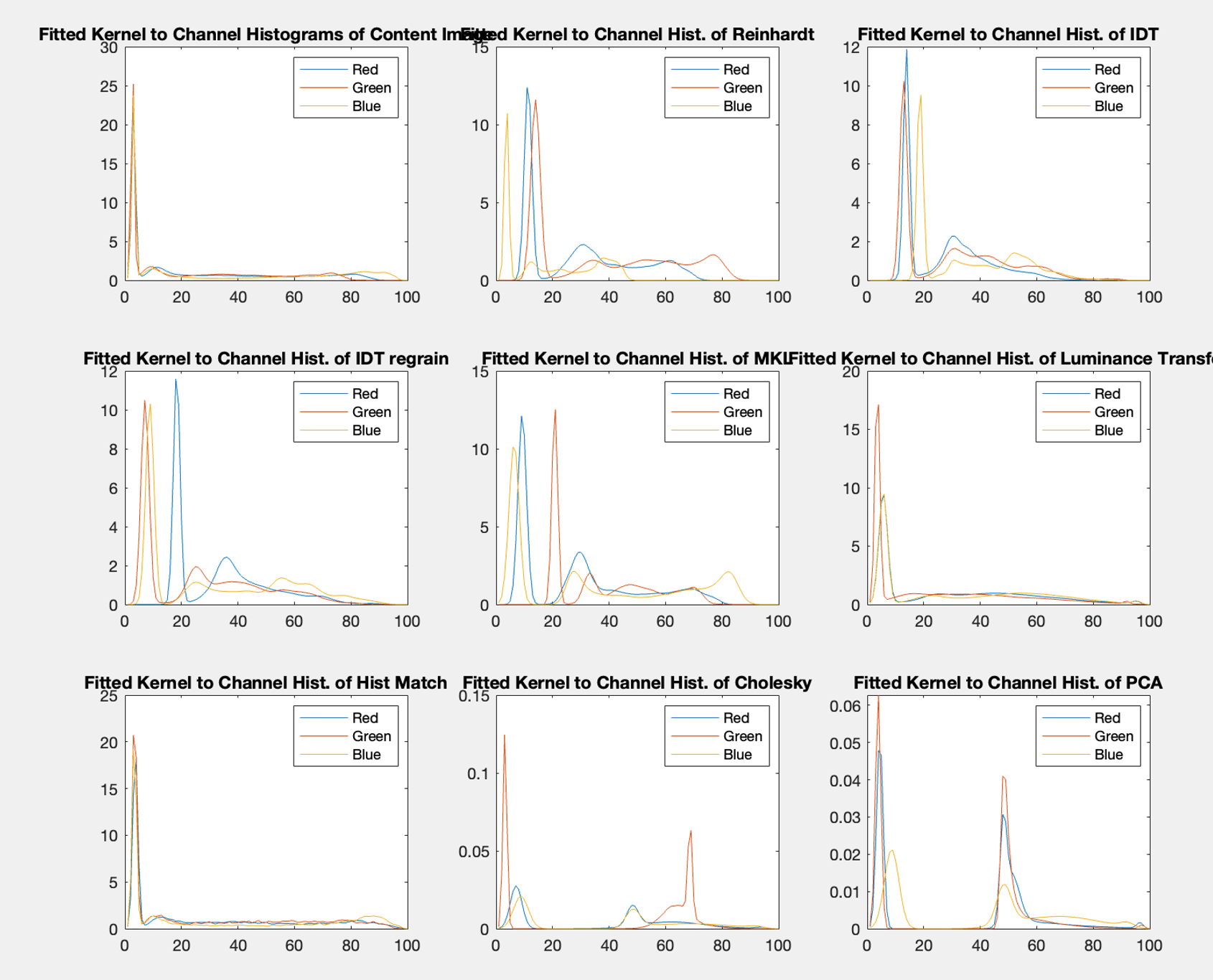} 
	\caption{Kullback-Leibler Divergence (Content to Style Transfer)} 
	\label{fig:bridgemonetkernel} 
\end{figure}

    Gatys et. al. notes that transferring the color distribution before a style transfer via neural training generates better visual quality results.  As shown in Figure \ref{fig:bridgehist}, Cholesky and the PAC transfers have a very small intensity dispersion and have one large sharp spike at a certain intensity and do not match the histograms of the content image.   The histograms for each of the color transfer algorithms are then converted into pdfs using kernel density estimation (\code{ksdensity} function in Matlab) as shown in Figure \ref{fig:bridgemonetkernel} so that the KL divergence between the pdf of the color channels in the content image and the generated style and color transfer images can be computed and compared.  
    
\begin{figure}[H]
    \centering
\includegraphics[width=0.8\columnwidth]{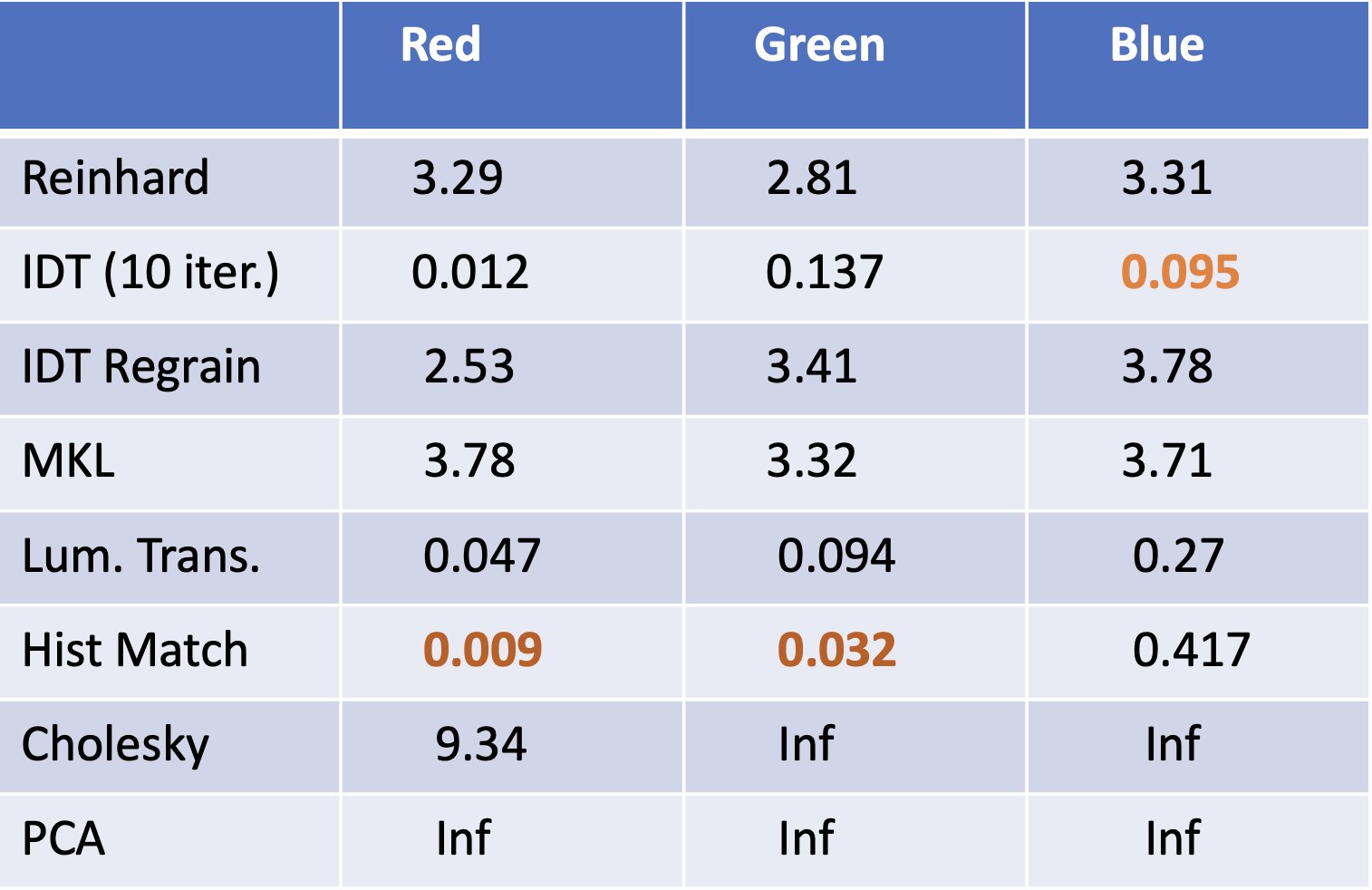} 
	\caption{Kullback-Leibler Divergence (Content to Style Transfer)} 
	\label{fig:bridgemonetkl} 
\end{figure}

    Figure \ref{fig:bridgemonetkl} shows the Kullback-Leibler Divergence for each of the kernel density estimated pdfs for each algorithm compared to the content image. 

\subsection{Experiment 2} 
\begin{figure}[H]
    \centering
	\includegraphics[width=0.9\columnwidth]{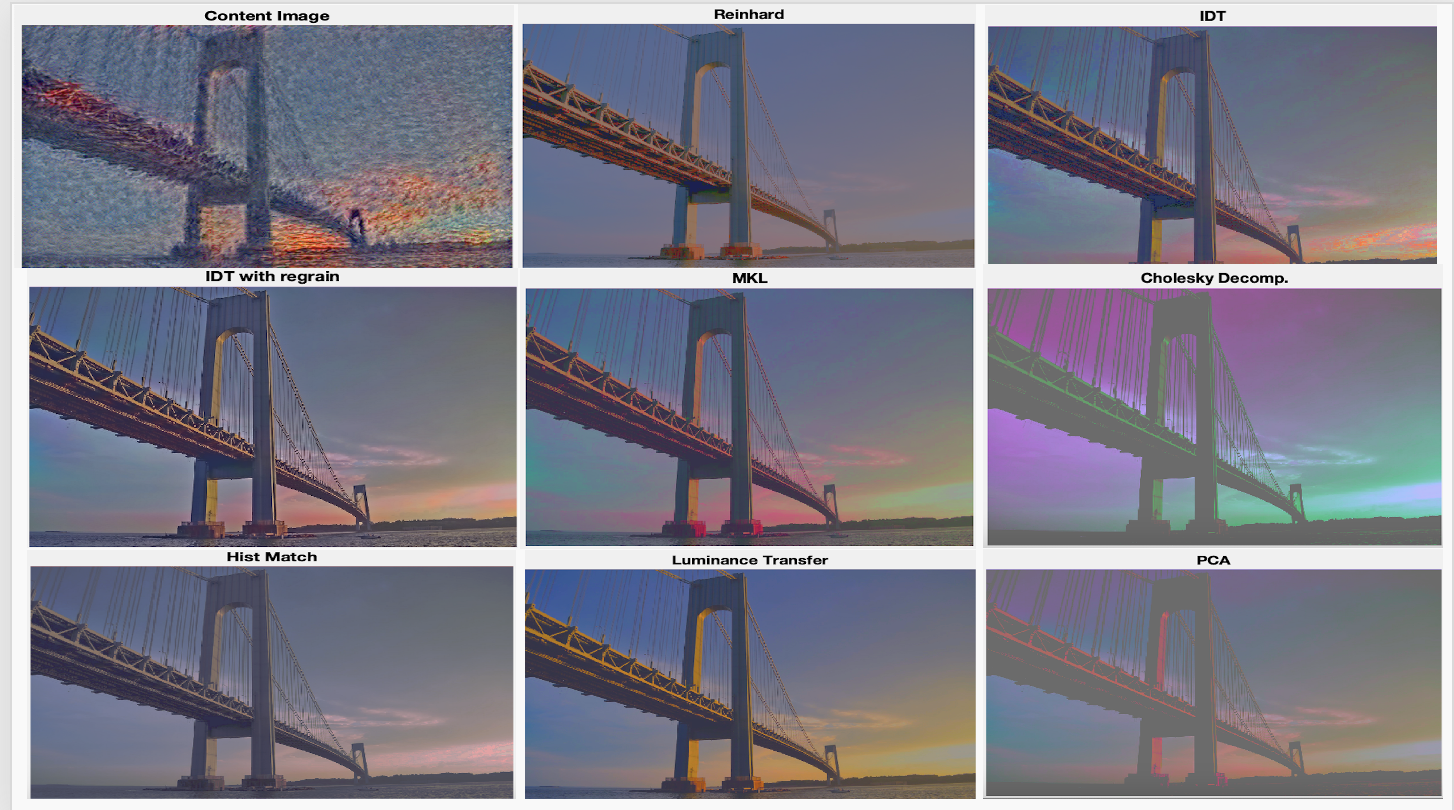} 
	\caption{Color Transfer (Style Color to Content Image Transfer)} 
	\label{fig:bridges4} 
\end{figure}

\begin{figure}[H]
    \centering
    \begin{subfigure}{0.4\textwidth}
    \includegraphics[width=\textwidth]{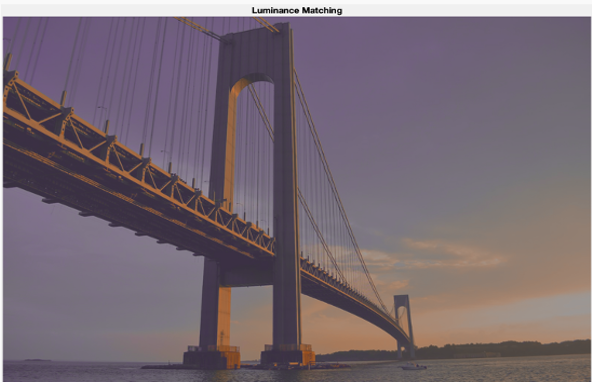} 
	\subcaption{Lum. Transfer without Lab conversion}
    \end{subfigure} 
    \begin{subfigure}{0.4\textwidth}
    \includegraphics[width=\textwidth]{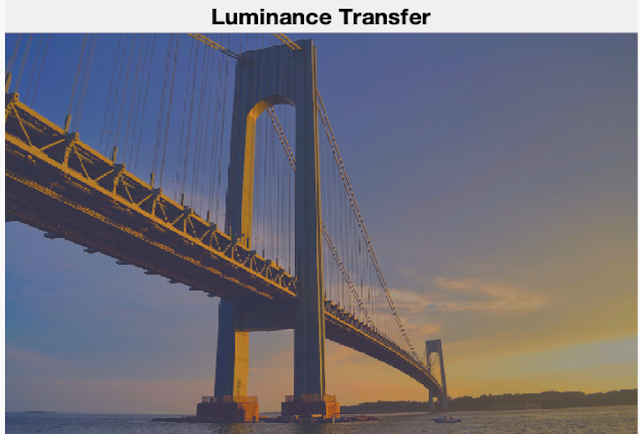} 
    \subcaption{Lum. Transfer with Lab conversion} 
    \end{subfigure} 
    \caption{Style Image to be Transferred to Content Image} 
	\label{fig:lab10} 
\end{figure}

    \ \ \ Figure \ref{fig:bridges4} shows the generated images using the color from the generated Money style image of the bridge in Experiment 1.  As shown, Reinhard, IDT, MKL, IDT with regrain, and histogram matching algorithms exhibit the color intensities and saturation levels in the generated style (source) image.  The luminance transfer provides the best sharpness and contrast, but has higher saturation than the other images.  The Cholesky decomposition does not transfer any of the correct colors and textures/details are lost.  On the other hand, though PCA transfers color intensities from the source image, but the image loses sharpness and details as the transferred colors are faded/muted and also have low saturation.
    Note that for certain algorithms like luminance transfer the color grading transfer will differ depending on whether the RGB image is first converted to logarithmic (Lab) color space first. Figure \ref{fig:lab10}(a) shows the luminance transfer image without the conversion and \ref{fig:lab10}(b) is with the conversion.  The image with Lab conversion gives a sharper and brigher image and better matches the color gradient of the generated style image.

\begin{figure}[H]
    \centering
	\includegraphics[width=0.9\columnwidth]{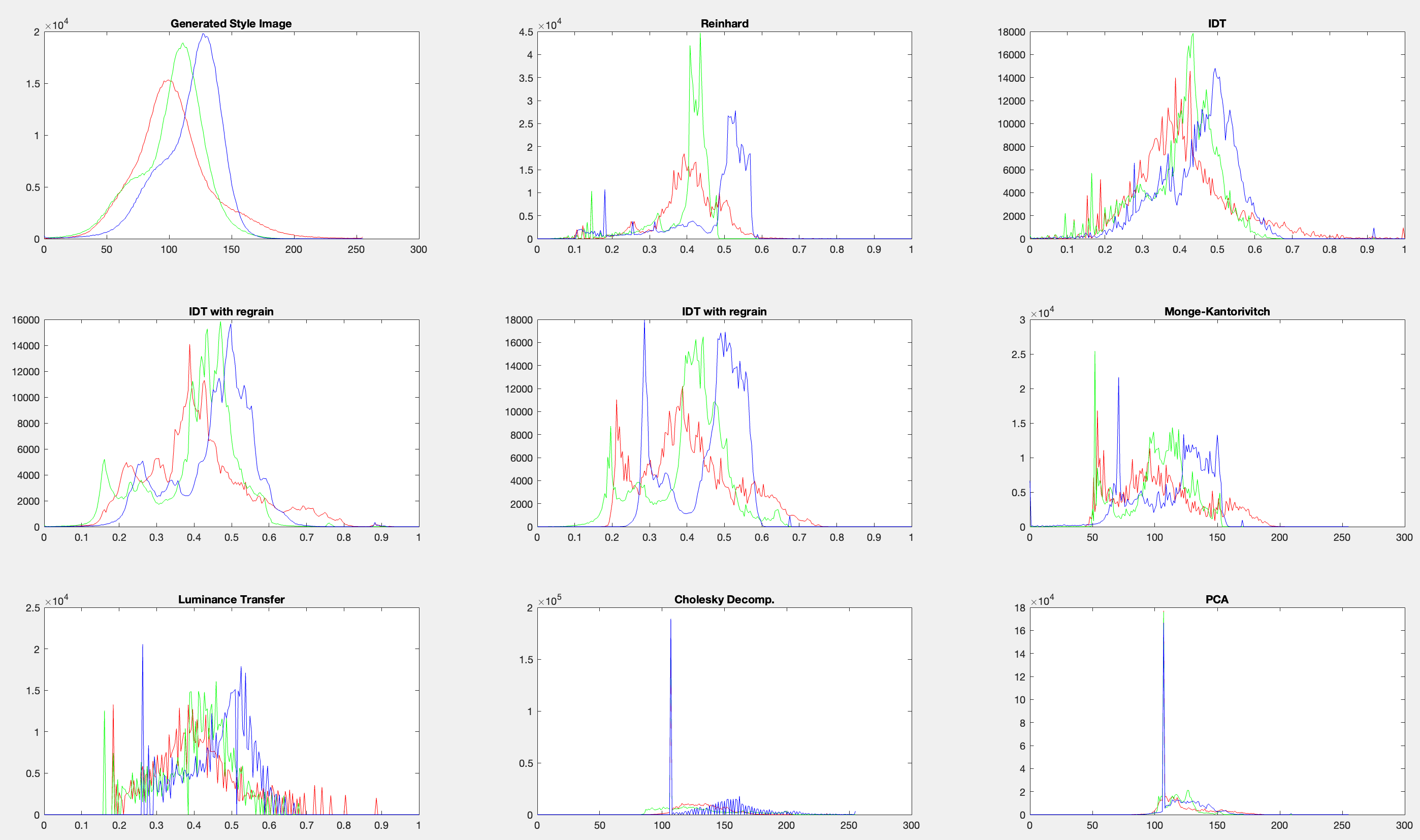} 
	\caption{Histograms (Style to Content Transfer)} 
	\label{fig:mshist} 
\end{figure}

\begin{figure}[H]
    \centering
	\includegraphics[width=0.9\columnwidth]{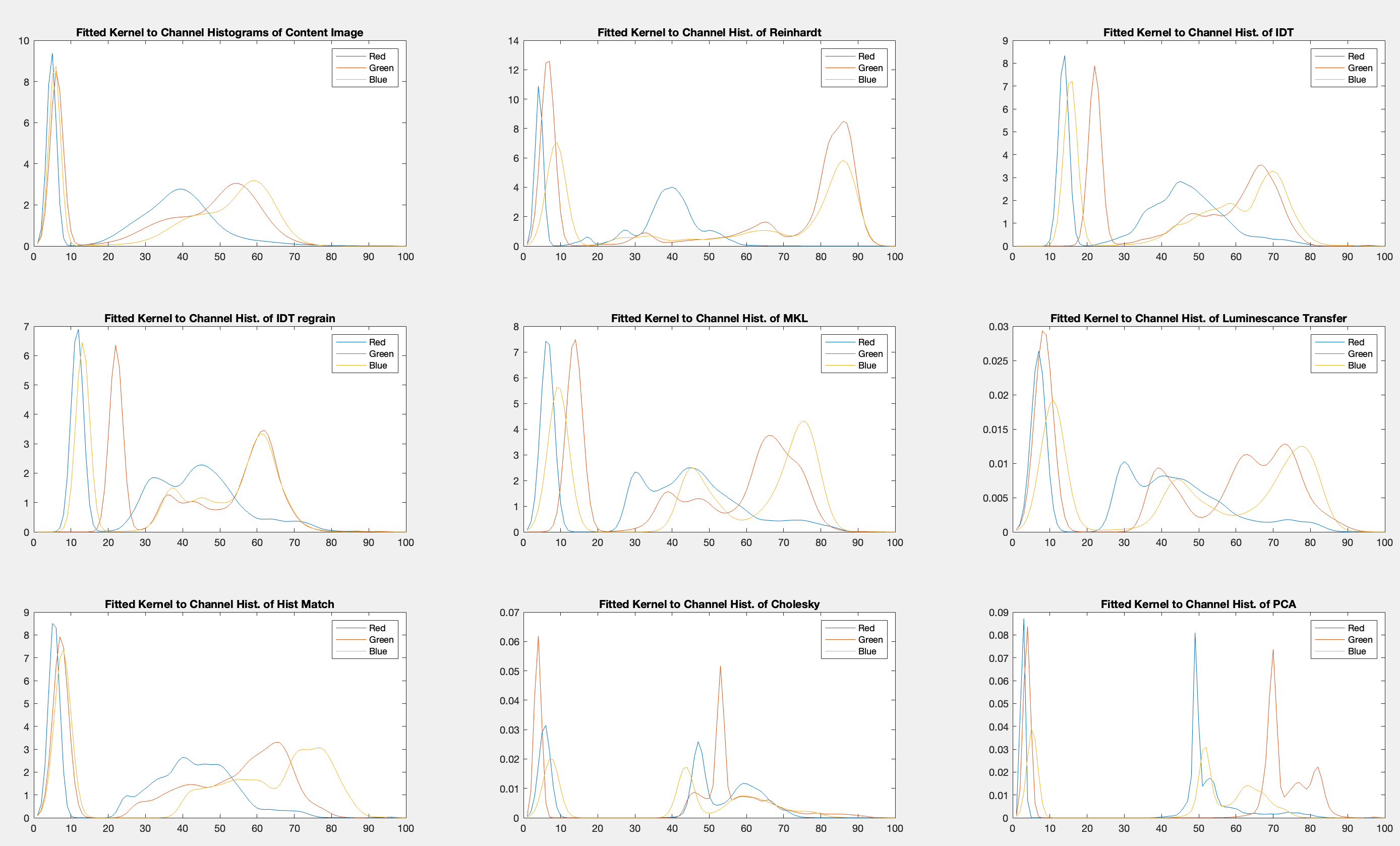} 
	\caption{Color Histograms}
	\label{fig:bridgedensityest} 
\end{figure}

\begin{figure}[H]
    \centering
	\includegraphics[width=0.7\columnwidth]{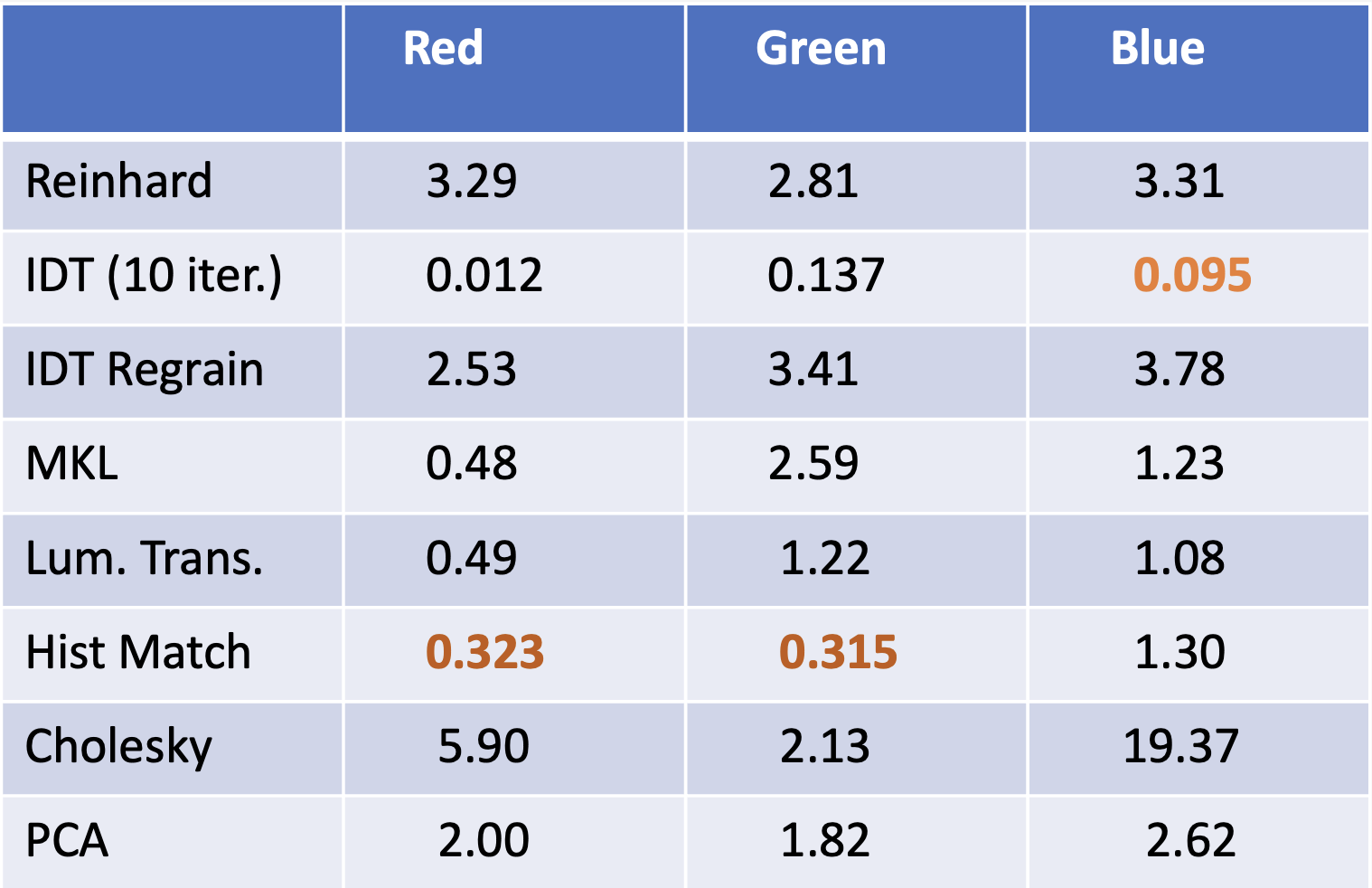} 
	\caption{Kullback-Leibler Divergence (Content to Style Transfer)} 
	\label{fig:monetkl} 
\end{figure}

    Figure \ref{fig:mshist} show the color histograms. The generated style image color histograms are smoother than those for the content source image in Experiment 1.   The other color histograms does not have is this smoothness but do have the intensity variation overlap.  The IDT and IDT with regrain seem to match the peaks and overlap of the generated style histograms better than the other algorithms.  As in Experiment 1, the Cholesky decomposition and PCA have a sharp spike and peak a certain blue channel intensity and poorly match the generated style histograms.   
    Figure \ref{fig:bridgedensityest} shows the fitted kernel densities to the color histograms.
    Figure \ref{fig:monetkl} shows the Kullback-Leibler Divergence for each of the kernel density estimated pdfs for each algorithm compared to the content image. 
   
\subsection{Experiment 3} 
\begin{figure}[H]
    \centering
    \begin{subfigure}{0.4\textwidth}
    \includegraphics[width=\textwidth]{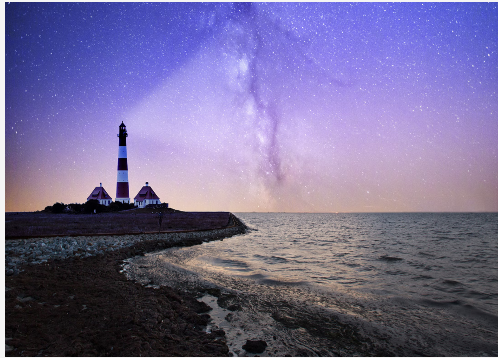} 
	\subcaption{Content Image}
    \end{subfigure} 
    \begin{subfigure}{0.4\textwidth}
    \includegraphics[width=\textwidth]{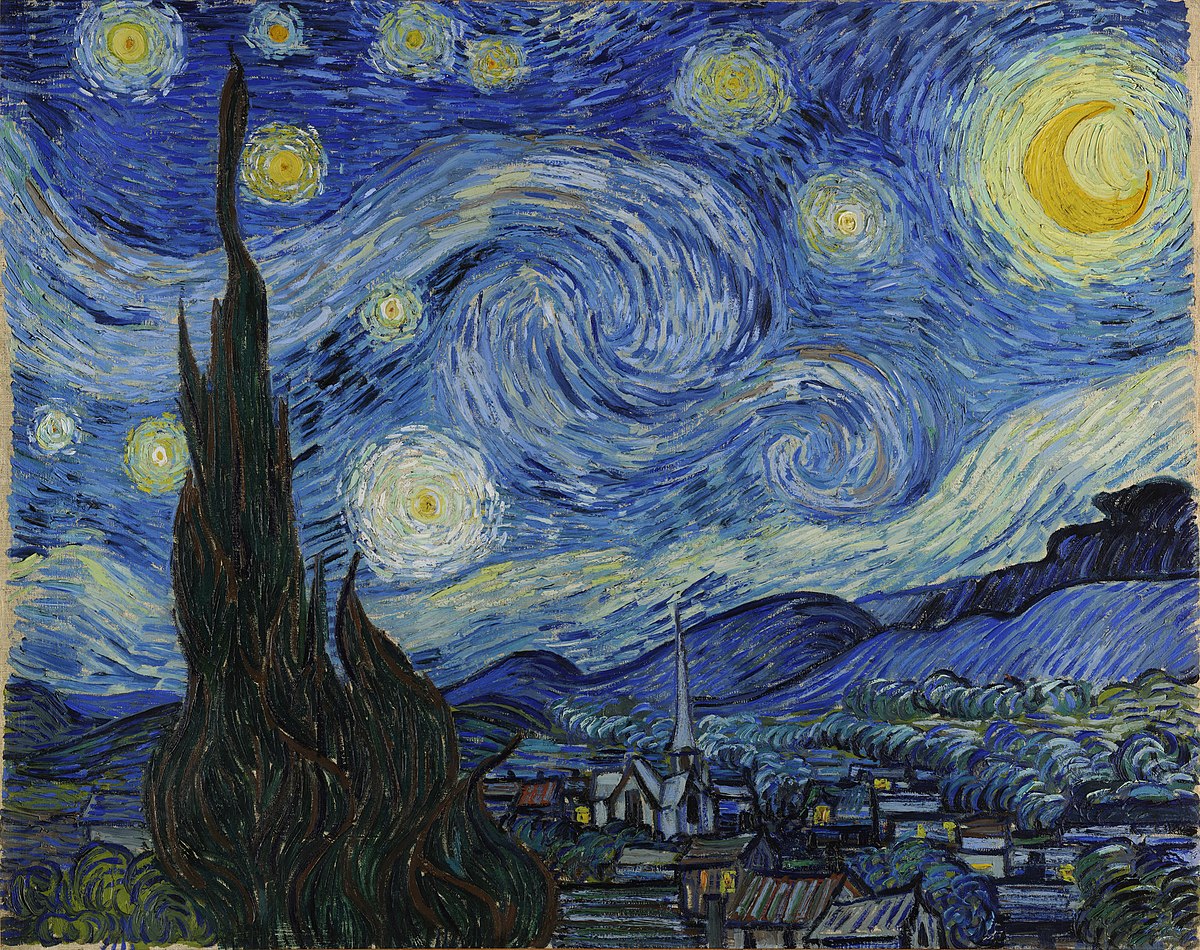} 
    \subcaption{Van Gogh Style Image} 
    \end{subfigure} 
    \caption{Style Image to be Transferred to Content Image} 
	\label{fig:style2} 
\end{figure}

\begin{figure}[H]
    \centering
	\includegraphics[width=0.9\columnwidth]{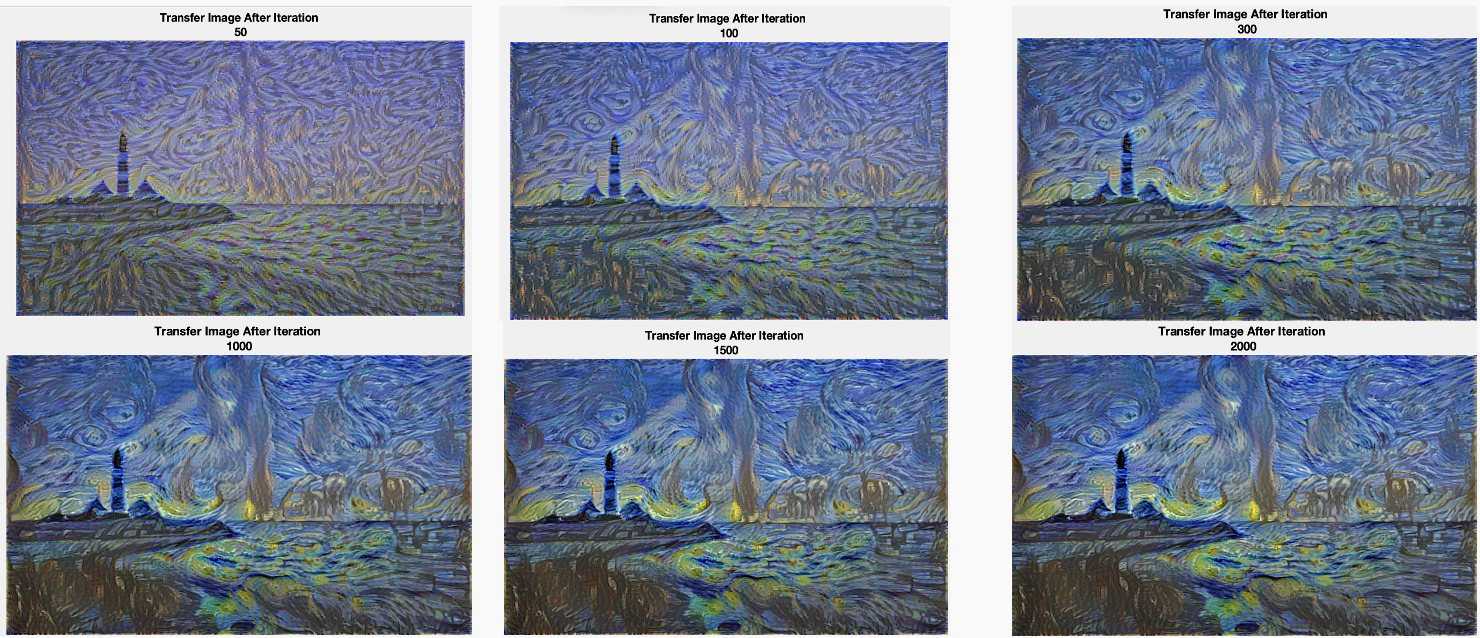} 
	\caption{Color transfer to Generated Van Gogh Style Image.}  
	\label{fig:lighthouses2}
\end{figure}

\begin{figure}[H]
    \centering
	\includegraphics[width=0.9\columnwidth]{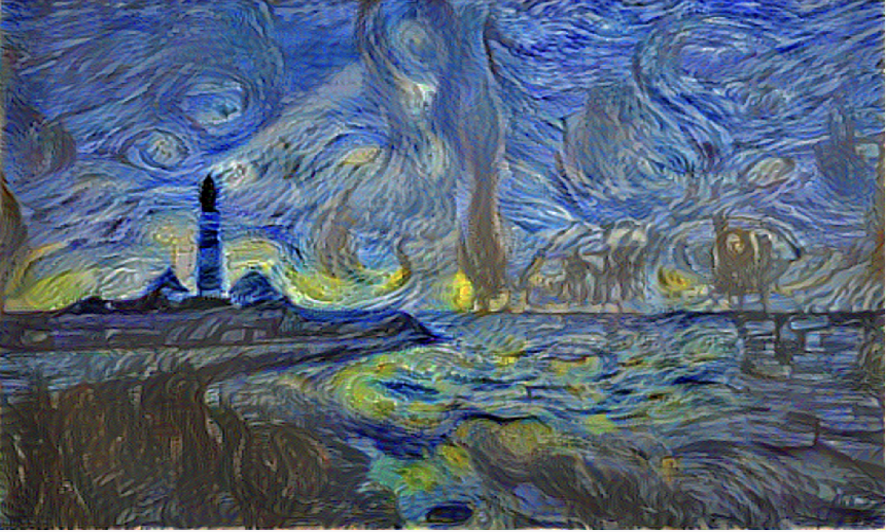} 
	\caption{Final generated style image}
	\label{fig:final2} 
\end{figure}

\begin{figure}[H]
    \centering
	\includegraphics[width=0.9\columnwidth]{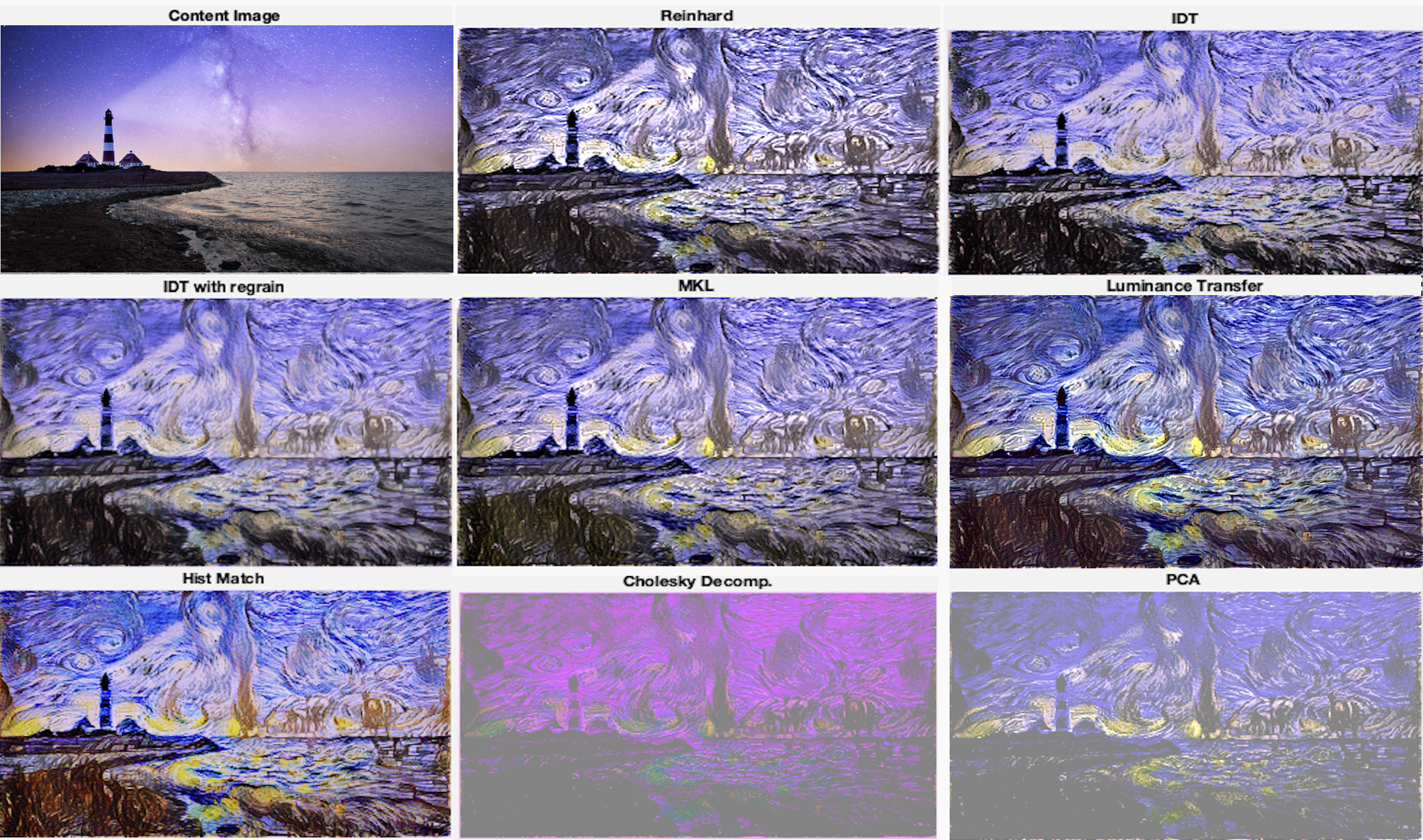} 
	\caption{Color Transfer Generated Images}
	\label{fig:ct2} 
\end{figure}

     \ \ \ The content image (lighthouse) and style image (Van Gogh's \textit{Starry Night}) are shown in Figure \ref{fig:style2}(a) and Figure \ref{fig:style2}(b), respectively.   The content image is $1234 \times 2122 \times 3$ and the style image is $1228 \times 2104 \times 3$.
    Figure \ref{fig:lighthouses2} shows the generated image at different iterations (1,100,300,100,1500,2000) in the neural training.  As shown the high-level (coarse) details are transferred in the early iterations and the more low-level (fine) details are transferred in the higher iterations.   
    The final generated style transfer image is shown in Figure \ref{fig:final2}.
    This image is then input into the different color transfer algorithms.  The color transfer generated images are shown in Figure \ref{fig:ct2}.
    The top left is the original source image.  The top middle is the Reinhard transfer.   The top right is the IDT transfer.   The middle row left image is IDT with regrain.   The middle row middle image has the color transfer using MKL.  The middle row right image has the color transfer using luminance transfer.   Finally, the bottom row left is the Cholesky transfer.  The middle row middle is histogram (equalization) matching and the middle left is the PCA.  

    As in experiment 1, the histogram matching provides the best perceptual-quality and contrast.   The intensity colors from the source image are all captured in the target image.  Though the luminace transfer does not appear to have as good perceptual quality and contrast as the IDL, IDL with regrain, and MKL, it suprisingly, has lower KL values for all of the channels than these algorithms. Since the luma component captures the brightness of the image.  The color histograms for each of the color transfer algorithms is shown in Figure \ref{fig:vangoughhist}:
 \begin{figure}[H]
    \centering \includegraphics[width=0.9\columnwidth]{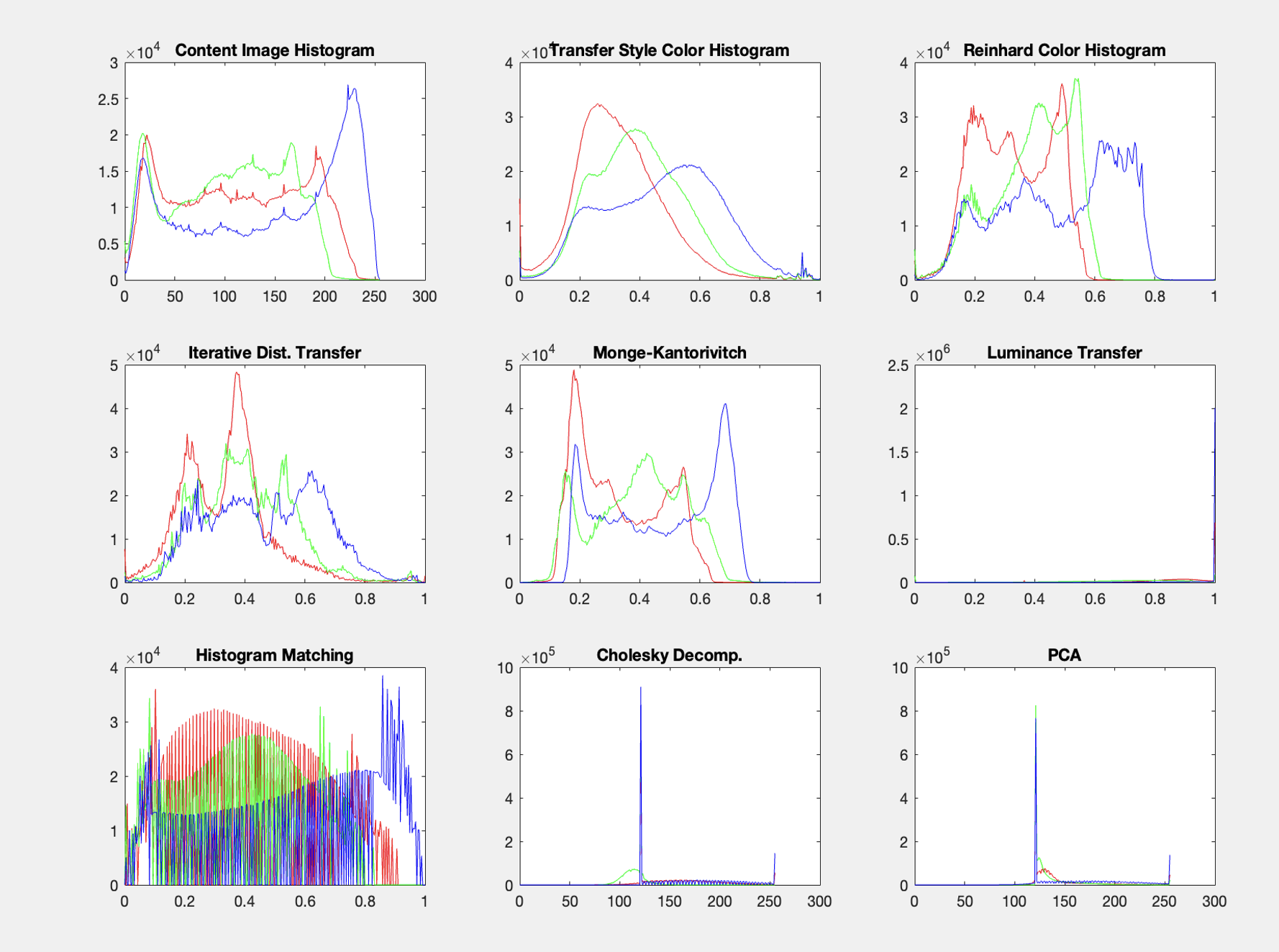} 
	\caption{Color Histograms}
	\label{fig:vangoughhist} 
\end{figure}      
    Ideally, we want the color transfer algorithm to have their color histograms match those of the content image (top left).  The Histogram Matching (lower left) matches the content histogram the closest of all the algorithms followed by the IDT.  This is expected because the histogram matching is actually \enquote{matching} the intensity distribution of the reference image via computation of the accumulated cdf.

    The Reinhard algorithm exhibits histogram shapes that have peaks like the content image histograms, but they  are not as well aligned with their position, locations, and the dispersion in the tails as the histogram matching algorithm.
    With luminance transfer, all the color channels follow a similar smooth bell-shaped curved shape and exhibit dispersion across the intensity values.  Though luminance transfer does not have the sharp peaks of the content image histograms, it is highlighting the luminance channel and thus is extracting and transferring the luminance channel from the content image.  
    
\begin{figure}[H]
    \centering
    \includegraphics[width=0.9\columnwidth]{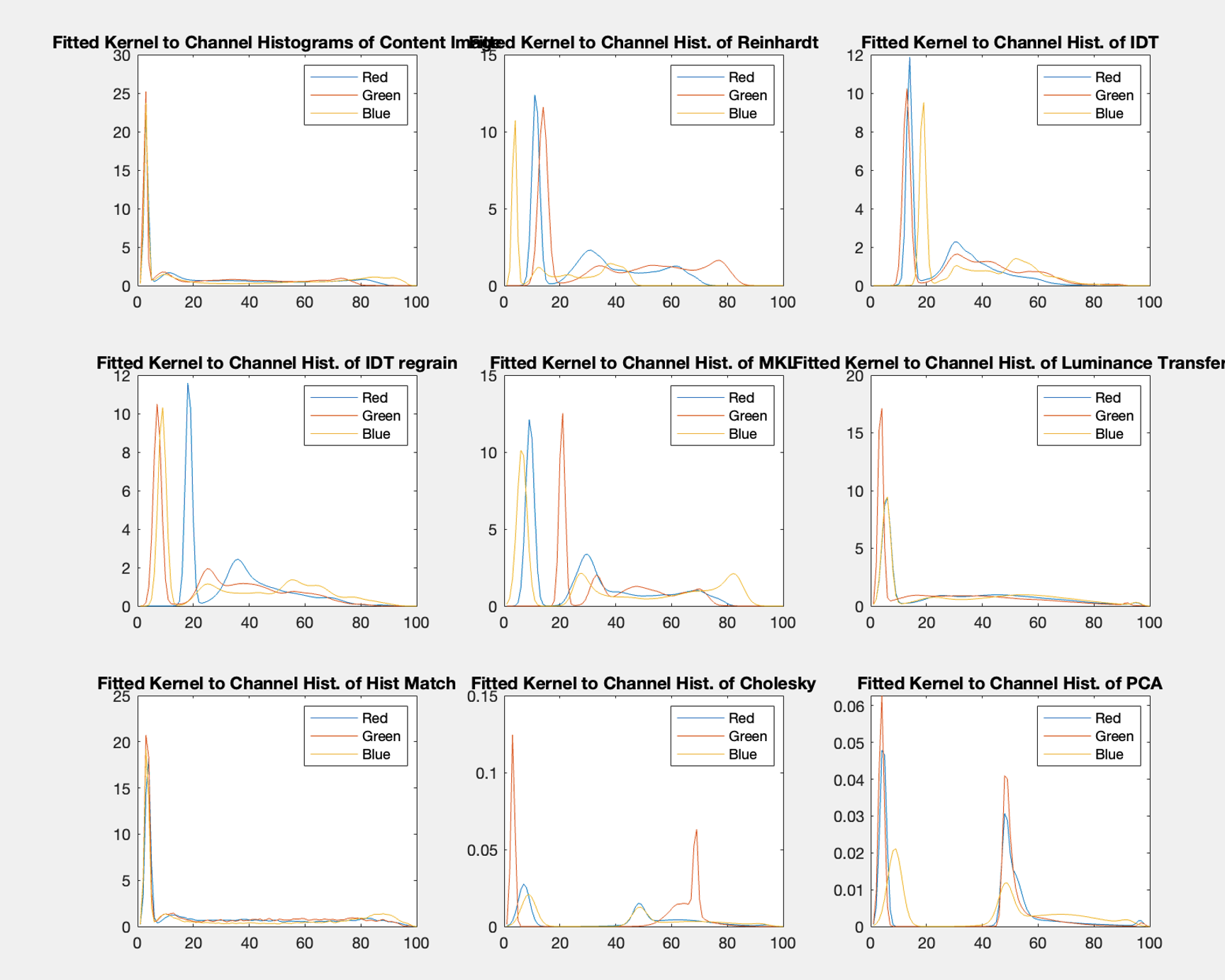} 
	\caption{Color Histograms}
	\label{fig:vgkernel} 
\end{figure}

    Gatys notes that performing luminance transfer from the original content image to the style image before training generates better visual quality.  The Cholesky and the PAC have very small intensity dispersions and have one large sharp spike at a certain intensity.  The histograms for each of the color transfer algorithms are then converted into pdfs using kernel density estimation (\code{ksdensity} function in Matlab) as shown in Figure \ref{fig:vgkernel} so that the KL divergence between the pdf of the color channels in the content image and the generated style and color transfer images can be computed.  

\begin{figure}[H]
    \centering
	\includegraphics[width=0.9\columnwidth]{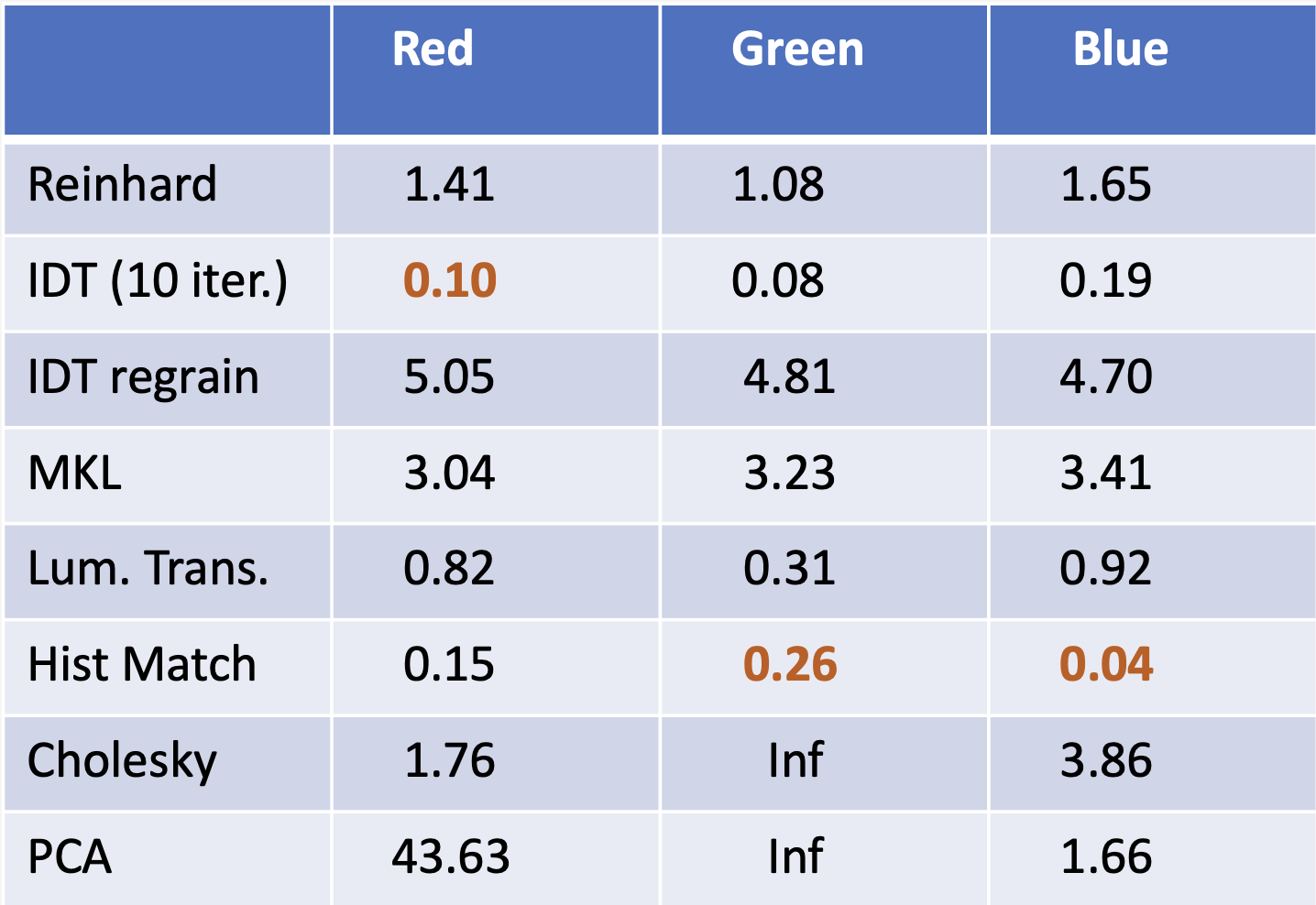} 
	\caption{Kullback-Leibler Divergence (Color Transfer Generated Images)} 
	\label{fig:vgkl} 
\end{figure}

    Figure \ref{fig:vgkl} shows the Kullback-Leibler Divergence for each of the kernel density estimated pdfs for the generated style image of each algorithm compared to the content image (lighthouse). 
    The histogram matching algorithm has the lowest KL for the green and blue channels, while the IDT has the lowest KL for the red channel.
   
\subsection{Experiment 4}
\begin{figure}[H]
    \centering
	\includegraphics[width=0.9\columnwidth]{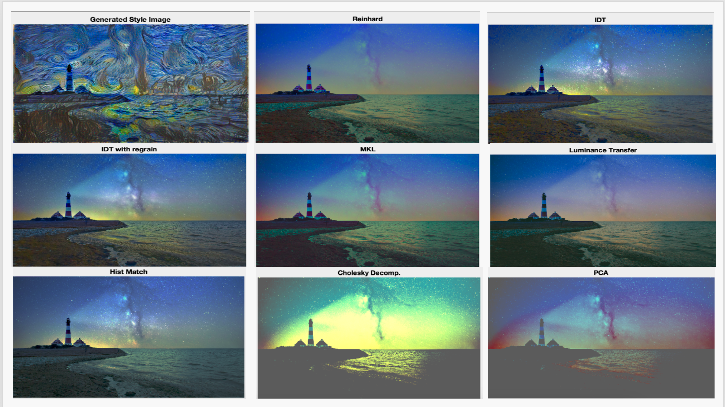} 
	\caption{Color Histograms}
	\label{fig:vgtransfer} 
\end{figure}

    \ \ \ For comparison purposes, we do a color transfer from the generated style image to the content image for the various algorithms as shown in Figure \ref{fig:vgtransfer}.
    The generated style image is in the top row left.   The Reinhard generated image is shown in the top middle.   The IDT generated image is in the top row right.   The IDT with regrain is shown in the middle row left.   The MKL is shown in the middle row center.  The luminance transfer is in the midde row right.    The histogram matching algorithm is shown in the lower row left.  The Cholesky decomposition is in the bottom row middle and the PCA is shown in the bottom row right.

    As shown, the color grading for the histogram matching, IDT and IDT with regrain is very similar to that the generated image.  The color grading for Reinhard, MKL, and luminance transfer are also good, but seem to include some pinkish-orange hues not in the generated image.  The Cholesky and PCA have very poor color grading matching.

\begin{figure}[H]
    \centering
	\includegraphics[width=0.9\columnwidth]{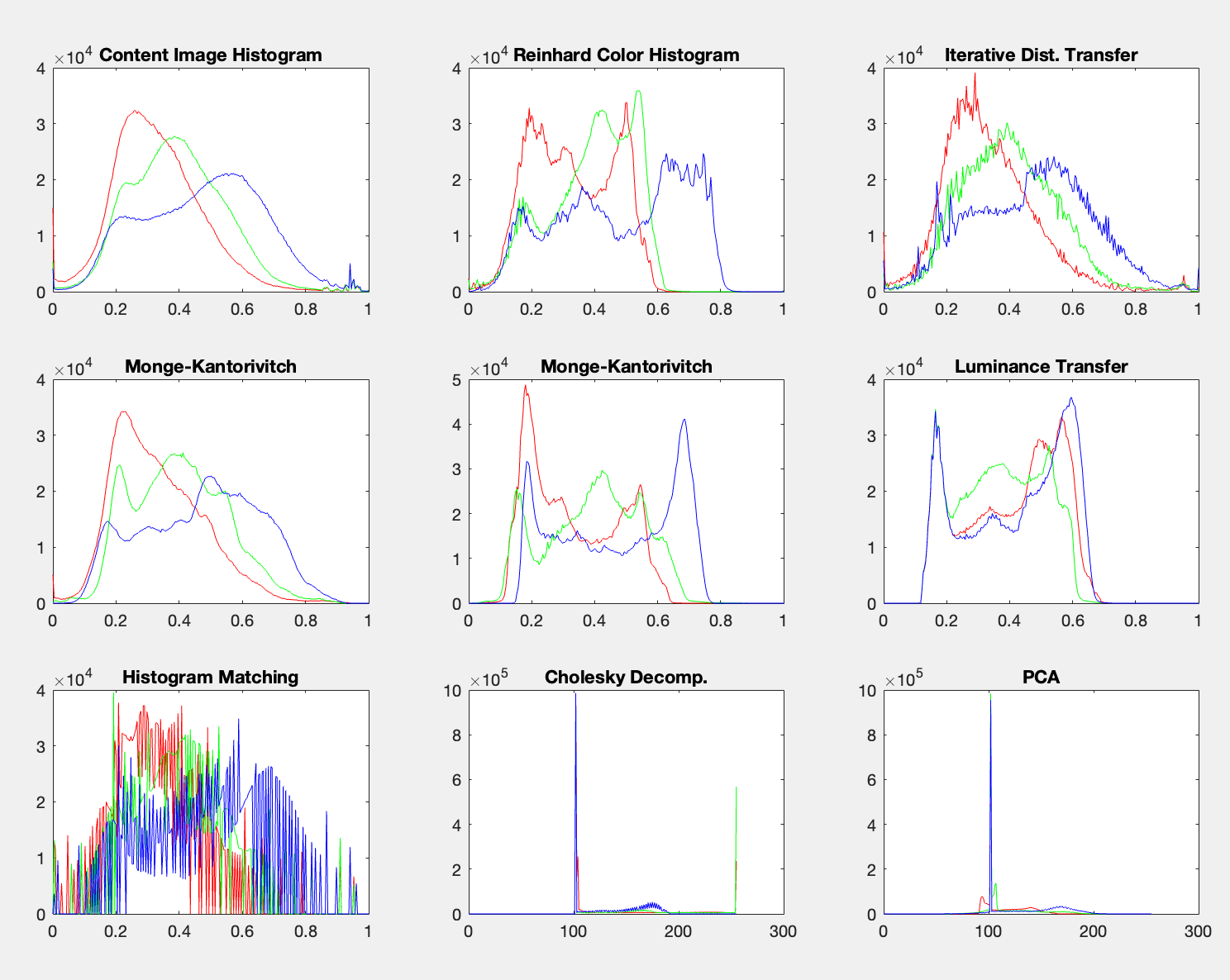} 
	\caption{Color Histograms (Style to Content Transfer)}
	\label{fig:vghist2} 
\end{figure}

\begin{figure}[H]
    \centering
	\includegraphics[width=0.9\columnwidth]{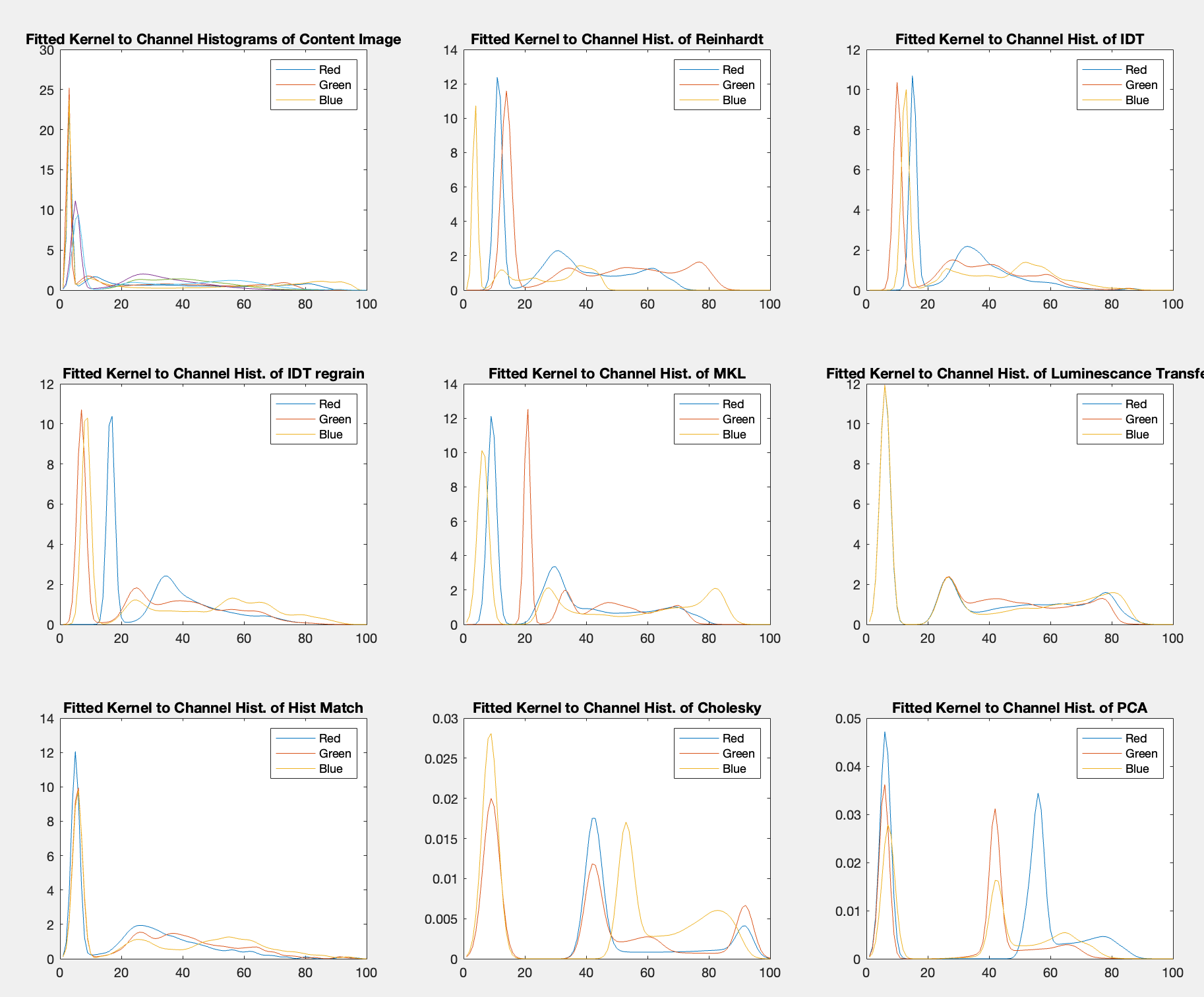} 
	\caption{Kernel Density Estimation (Style to Content Transfer)}
	\label{fig:vgtransferkernel} 
\end{figure}

    The color channel histograms for the algorithms are shown in Figure \ref{fig:vghist2} 
    Figure \ref{fig:vgtransferkernel} show the kernel density estimates of the color channel histograms for each of the algorithms.

\begin{figure}[H]
    \centering
	\includegraphics[width=0.9\columnwidth]{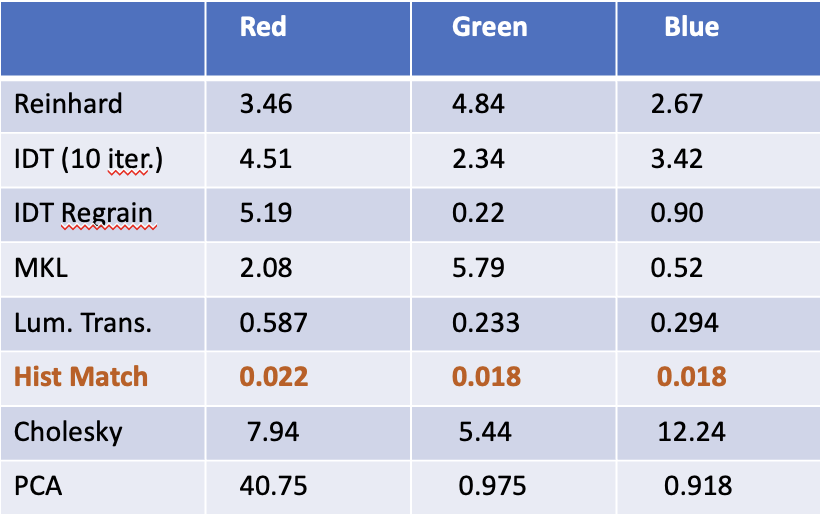} 
	\caption{Kullback-Leibler Divergence (Style to Content Transfer)} 
	\label{fig:vgtransfer4} 
\end{figure}

    Figure \ref{fig:vgtransfer4} shows the Kullback-Leibler Divergence for each of the kernel density estimated pdfs for each color channel of the histograms of the images generated by each algorithm compared to the generated style image (upper left hand corner of Figure \ref{fig:final2}. 

\section{Analysis}

    \ \ \ \ \ \ As measured by having very low (near zero) KL values for all images, histogram matching equalization generally outperforms all other algorithms for all experiments.  This can be explained by the direct computation of computing intensity probabilities and generating the cdf for each channel to match that of the source content image across all intensity levels.   However, the color distribution often cannot be matched identically, leading to a color mismatch between the colors of the generated output image and the content image \cite{Gatys:2016}.  

    Luminance transfer provides the next best color transfer performance.  One should convert the image into logarithmic color space before performing the luminance transfer as one will get much better results in terms of the perceptual-quality and lower KL values.  After the luminance transfer is completed, then one should convert back to RGB space. 
    
    In contrast to a square root decomposition, the Cholesky method does not perform well from a visual and perceptual-quality perspective.  This because it does not maintain the property of independence to orthogonal transformations.   Having such a property is necessary so that a color transfer has the same effect under different color spaces. Moreover, the Cholesky method \enquote{has the conceptually-undesirable property that it depends on the channel ordering, i.e. using RGB images will give different results from BGR. \cite{Gatys:2016}} 
    
    Converting from RGB to Logarithmic Lab color space prior to performing Cholesky and PCA eliminates infinity values for the KL values and in fact substantially reduces them to the values comparable to the other algorithms. However, for unknown reasons which need to be further explored, when converting back to RGB and casting with uint8, the generated images are completely black so one can not view the color transfer.  
    
    Thus, it is evident that the Cholesky and PCA can perform well based on the low KL values when the conversion is used, but given the importance of being able to view the actual color transfer image, it was decided to not do the Lab conversion beforehand at the expense of having extremely poor KL results. 

\section{Conclusion} 

    \ \ \ Color transfer algorithms are an important in digital image processing by adjusting the color information in a target image based on the colors in the source image.   They ensure the color gradient has a smooth and harmonious transition.  Color transfer enhances images and videos in film and photography, and can aid in image correction.

    The experiments in this project contributed to the literature by providing a better understanding of the type of results that color transfer algorithms and in particular, measuring how well they transfer to the generated target image both from a visual perceptual quality viewpoint and in relation to the color map in the source image.  None of the papers in the references actually measure the Kullback-Leibler divergence across different algorithms. Unlike most of the references, this project also illustrated the use of both style and color transfer algorithms as opposed to just one or the other.

    The experiments performed demonstrate that histogram matching, luminance transfer, IDT, IDT with regrain, and Kantorovitch
    are very effective color transfer algorithms for providing color transfer fidelity while Cholesky and PCA are not.  In general, histogram matching outperforms other algorithms because it ensures equal representation of the different pixel intensities in the source image are transferred to the generated target image. 

    However, though the experiments indicate that while histogram matching typically has the best KL performance as measured by having the lowest entropy, visual perceptual differences in the color grading (saturation, hue, and luminance) transferred to the generated image match closer to the source image and be preferred.  For instance, the luminance transfer, IDT, or IDT with regrain have color grading that appears to better match the color grading in the content image in Experiment 1.  Luminance transfer generates yields sharp quality and bright target images such in Experiment 2 and is a simple algorithm.

    Further research is required to understand the computational times and expense for automating these algorithms for use to many film frames rather than static images.  Additional experiments are also needed such as whether the transfers are improved before or after neural style transfer training and how they perform using different deep CNNs with different network structures such as ResNet, InceptionV3, and EfficientNet.  
    
    Future work is needed to better understand not only the impact of using different networks, but also on how parameter fine-tuning such as the number of training epochs and layer weights impact style and color transfer and whether and the degree that color transfer can be influenced by style.  The current training only used style and content loss functions and future work should also include total variational loss.  In addition, analysis of local transfer methods should be performed and compared to the global methods used in this project.

    The results are limited to only the images used in this project.  Numerous other experiments need to be performed with different style and quality of images (e.g. low and high resolution images, and the impact that image filtering/sharpening may also have on the results, to be able to generalize the results for applications in digital image processing.  Finally, the KL results were based on general kernel density fitting estimation as opposed to the KL Epanechnikov kernel approach suggested by Piti$\acute{e}$ \cite{Pitie:2007} in equation \ref{eq:epanech} due to lack of details on proper implementation of the variable scale.
    
\newpage 

\section{Appendix}

\subsection{Image Analogies} 

\ \ \ Image Analogies \cite{Hertzmann:2001} develops a framework which can create new images from a single training example and transferring the style to the new generated image.  
 
    This is accomplished through the idea of an image analogy: given an image $A$ and its transformation $A’$, provide any image B to generate an output image $B’$ that is analogous to $A’$.  More succinctly: $A : A’ :: B : B’$.  

    In the algorithm, the RBG channels of $A(p)$, $A'(p)$, and $B(q)$ are inputs, where $A$ is the unfiltered source image, $A'$ is the filtered source image, and $B$ is the unfiltered target image $B$.  The algorithm produces a filtered target image $B'$ as output as illustrated in Figure \ref{fig:analogy}:   
	\begin{figure}[H]
    \centering
	\includegraphics[width=0.8\columnwidth]{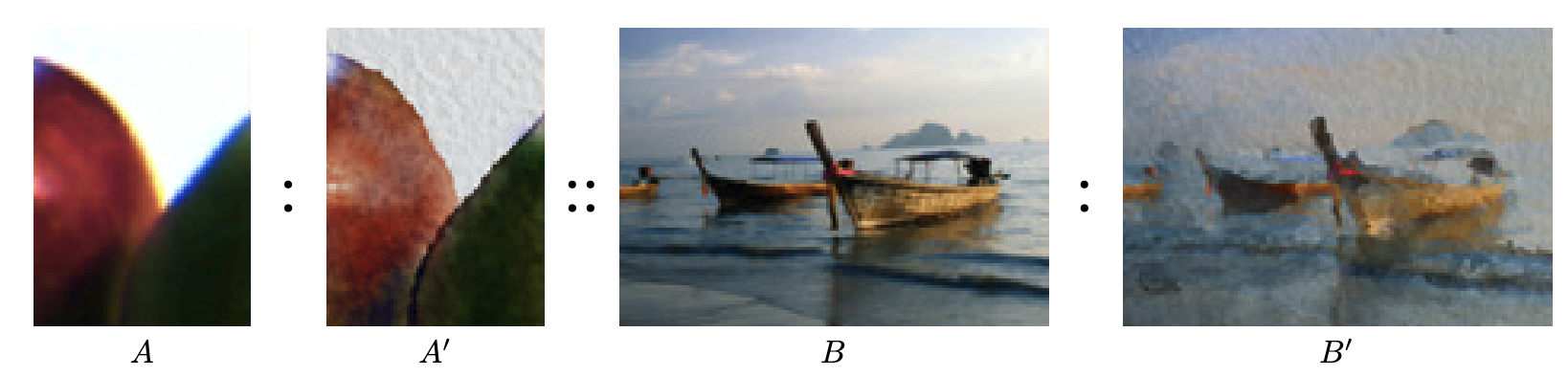} 
	\caption{Illustration of image analogies.  \cite{Hertzmann:2001}}
	\label{fig:analogy} 
    \end{figure} 
    
    The method assumes that the colors at and around any give pixel $p$ in $A$ correspond to the colors at and around that same pixel $p$ in $A'$, through the image filter that we seek to learn.

    In the initialization phase, multiscale Gaussian pyramid representations of $A, A'$, and $B$ are constructed along with their feature vectors and some additional indices used for speeding the matching process such as an approximate-nearest-neighbor search (ANN).  
    
    Following Hertzmann et al. (2001), \enquote{the synthesis proceeds from coarsest resolution to finest, computing multiscale representations of $B'$, one level at a time.  At each level $\mathcal{l}$, statistics pertaining to each pixel $q$ in the target pair are compared against statistics for every pixel $p$ in the source pair, and the 'best' match is found.    The feature vector $B'_{\mathcal{l}}(q)$ is then set to the feature vector $A'_{\mathcal{l}}(p)$ for the closest-matching pixel $p$, and the pixel that matched beet is recorded in $s_{l}(q)$ \cite{Hertzmann:2001}}.  

    The $\code{CreateImageAnalogy}$ algorithm is provided below in Figure \ref{fig:ia}.
  \begin{figure}[H]
    \centering
	\includegraphics[width=0.8\columnwidth]{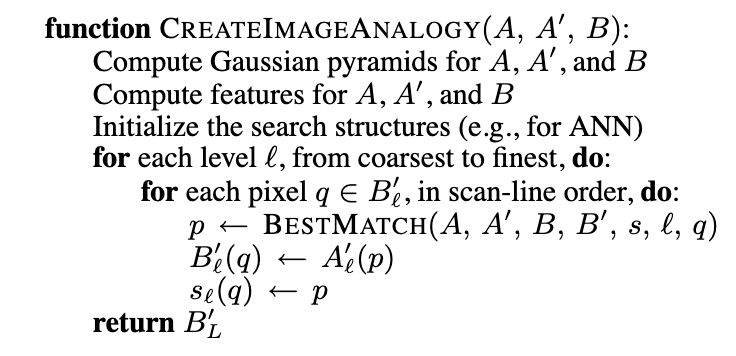} 
	\caption{$\code{CreatImageAnalogy}$ Algorithm. \cite{Hertzmann:2001}} 
	\label{fig:ia} 
\end{figure}    
    The $\code{BestMatch}$ algorithm referenced in the $\code{CreateImageAnalogy}$ is given below in Figure \ref{fig:ia2}:
\begin{figure}[H]
    \centering
	\includegraphics[width=0.8\columnwidth]{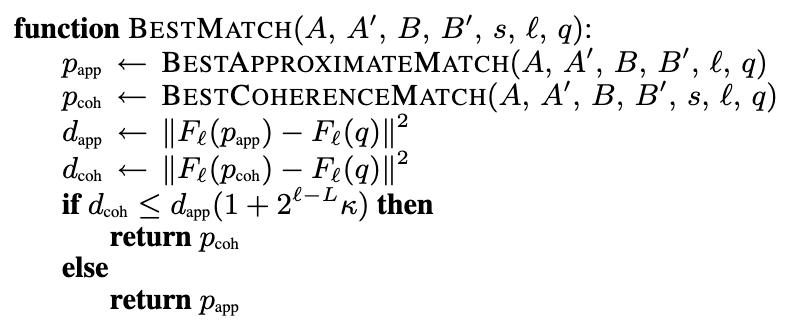} 
	\caption{\code{BestMatch} algorithm.  \cite{Hertzmann:2001}} 
	\label{fig:ia2} 
\end{figure}   
	Hertzmann’s algorithm attempts to learn complex, non-linear transformations between $A$ and $A’$, allowing them to be applied to any other image $B'$.    The general idea of image analogies is to, for each layer in a Gaussian pyramid, to iterate over each pixel, $q$, in $B'$ and find some pixel from A' that is the best match to $q$.  The relevant features from $p$ are then copies over onto $B'$, and the finest $B'$ layer from the pyramid is then returned as illustrated in Figure \ref{fig:analogy}.\footnote{See  \url{https://github.com/jmecom/image-analogies/} which uses another C++ ANN wrapper class \url{https://github.com/jefferislab/MatlabSupport/tree/master/ann_wrapper}.} 

    Gatys et al. found that the color matching works reasonably well with neural style transfer unlike Image Analogies which gave poor synthesis results.\footnote{\cite{Gatys:2016}, p.3.  A deep learning implementation of Image Analogies can be found at \url{https://github.com/msracver/Deep-Image-Analogy} \cite{Liao:2017}.}

\newpage
\printbibliography
\end{document}